\PassOptionsToPackage{dvipsnames}{xcolor}
\documentclass[singlecolumn]{NTUAI4XPaperEN}

\usepackage{lipsum}       
\usepackage{siunitx}      
\usepackage{amsfonts}     
\usepackage{mathtools}    
\usepackage{booktabs}     
\usepackage{listings}     
\usepackage{algorithm}
\usepackage{algpseudocode}
\usepackage{tikz}
\usetikzlibrary{positioning,arrows.meta}

\usepackage{nicefrac}       
\usepackage{microtype}      

\usepackage{graphicx}
\usepackage{subcaption} 

\usepackage{amsmath}
\usepackage{amsthm}
\usepackage{amssymb}

\usepackage{pifont}
\usepackage{ulem}

\usepackage{float}
\usepackage{algorithm}

\usepackage{multirow}
\usepackage{adjustbox}
\usepackage{amssymb}
\usepackage{multirow}
\usepackage{mdframed}
\usepackage{caption}
\usepackage{makecell}
\usepackage{graphicx}
\usepackage{algpseudocode}
\usepackage{algorithmicx}
\usepackage{wrapfig}
\usepackage{titletoc}

\crefname{section}{\S}{\S}
\crefname{table}{Tab.}{Tabs.}
\Crefname{table}{Table}{Tables}
\crefname{figure}{Fig.}{Figs.}

\titlecontents{section}
  [2em]
  {\bfseries}
  {\contentslabel{1.8em}}
  {}
  {\titlerule*[0.8pc]{.}\contentspage}

\titlecontents{subsection}
  [4.5em]
  {}
  {\contentslabel{3em}}
  {}
  {\titlerule*[0.8pc]{.}\contentspage}

\newcommand{\mcnl}{\\}
\newcommand{\increase}[1]{\sffamily\textcolor{green!55!black}{#1}}
\newcommand{\decrease}[1]{\sffamily\textcolor{red!65!black}{#1}}
\newcommand{\meanvalue}[1]{\sffamily\textcolor{black}{#1}}

\definecolor{Light}{rgb}{0.92, 0.99, 0.95}
\definecolor{mygray}{HTML}{E6E6FA}

\usetikzlibrary{positioning,arrows.meta}

\setthemecolor{1a2a6c}

\settitlepreskip{-10ex}
\settitlelogogap{3em}

\titlelogocontent{%
	\begin{minipage}{\textwidth}
		\begin{minipage}[c]{0.35\textwidth}
			\raggedright
			\includegraphics[height=1cm]{figs/ntu_ai4x_logo_merge_1.png}
		\end{minipage}%
		\hfill
		\begin{minipage}[c]{0.60\textwidth}
			\raggedleft
			\includegraphics[height=1cm]{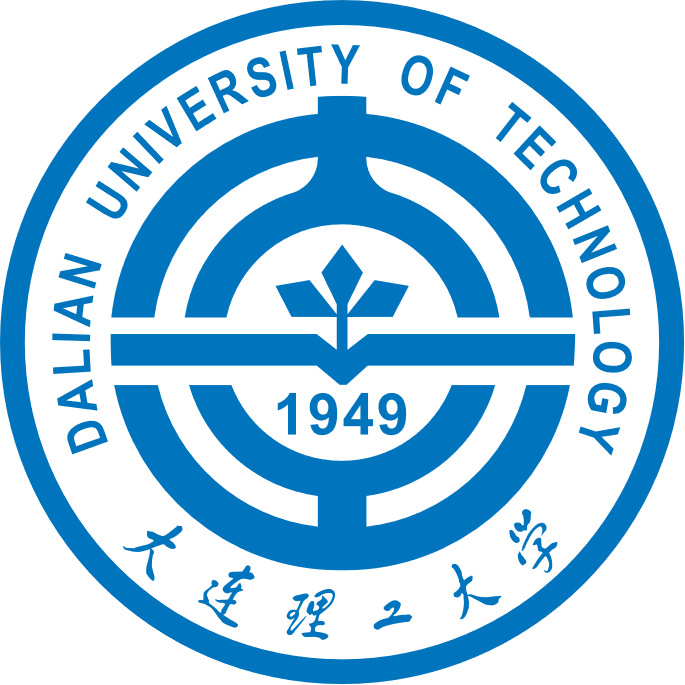}\hspace{0.0em}
            \includegraphics[height=1cm]{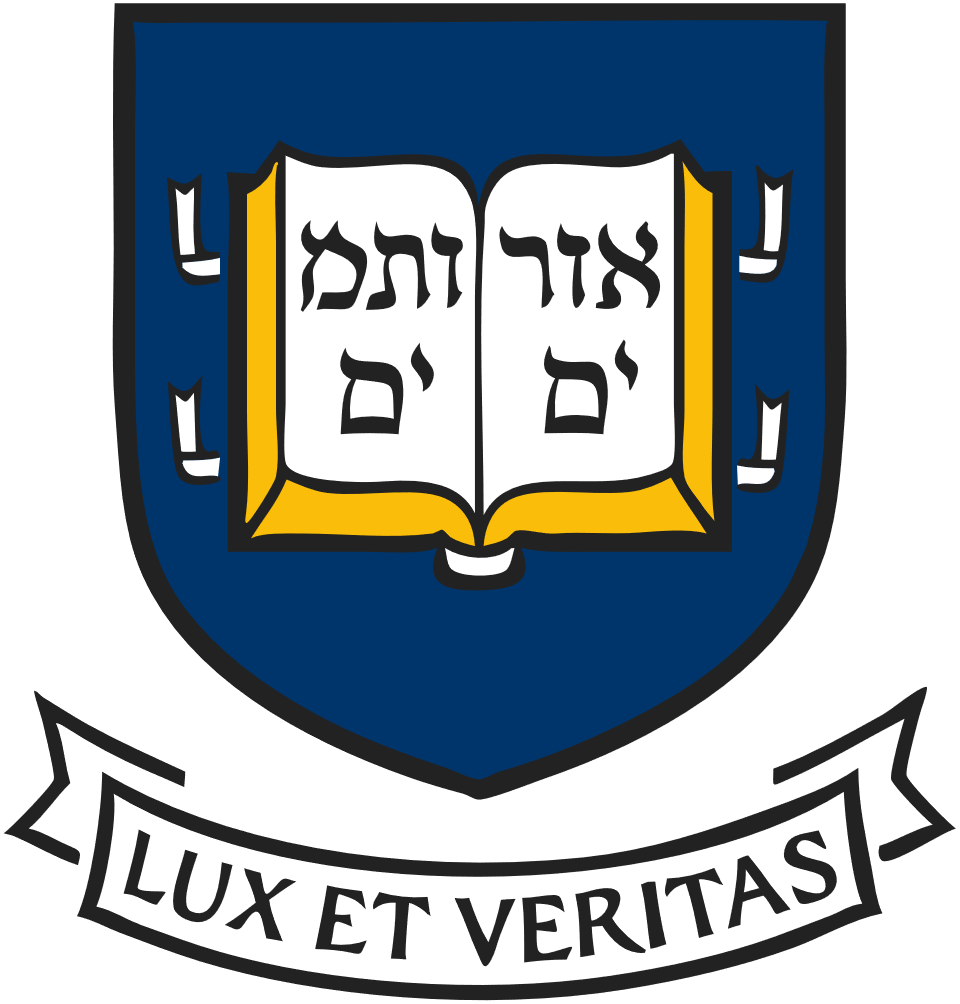}\hspace{0.0em}
            \includegraphics[height=1cm]{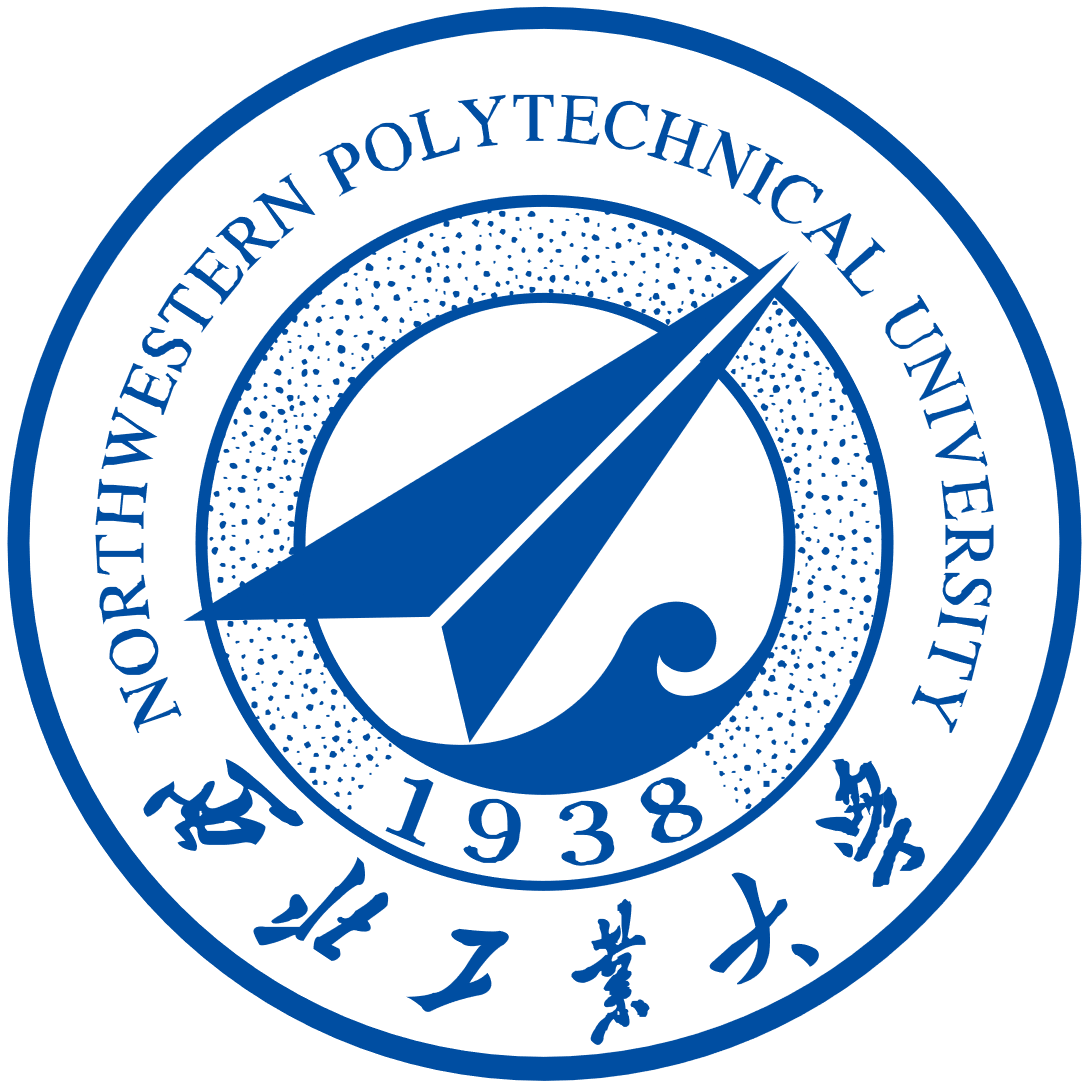}\hspace{0.0em}
              \includegraphics[height=1cm]{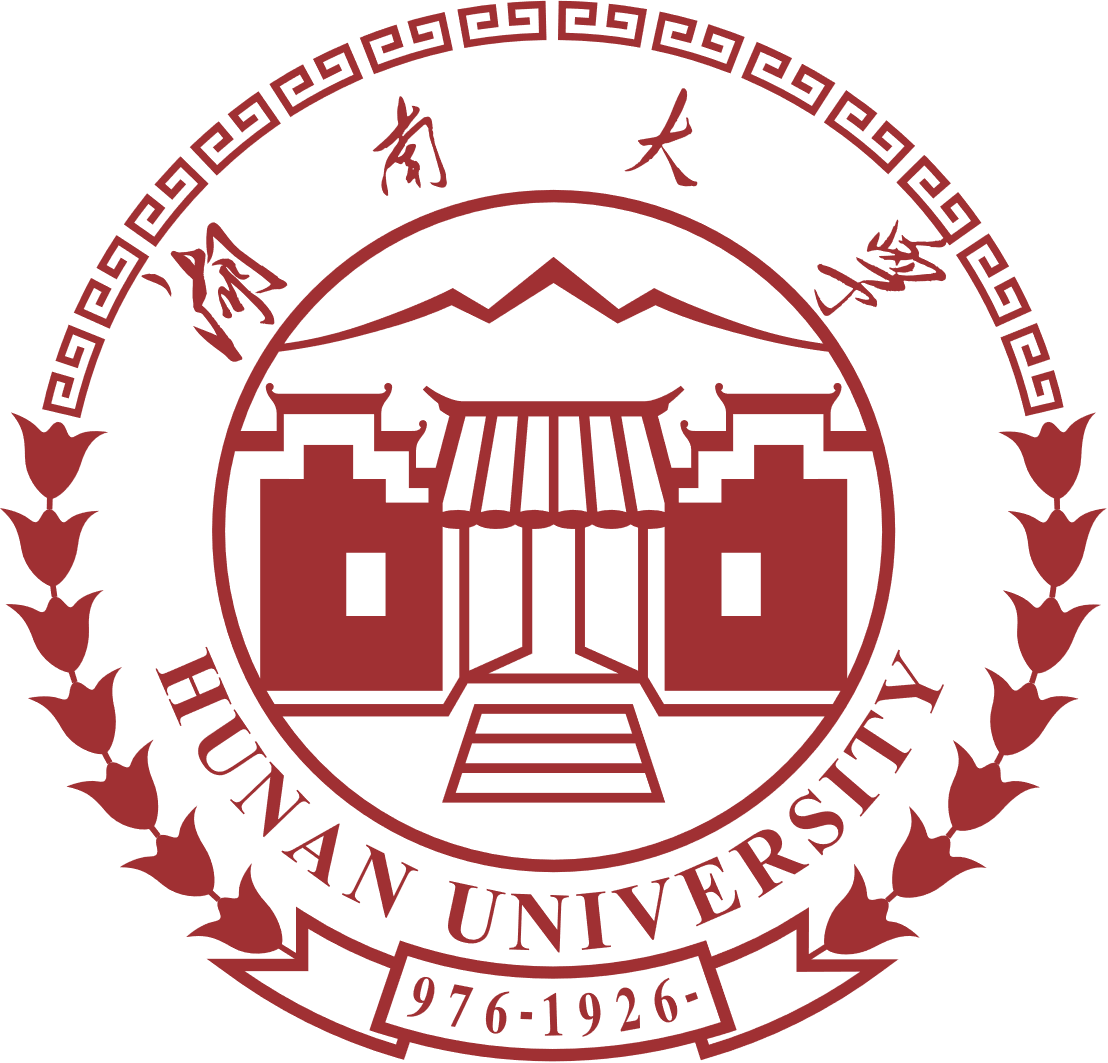}\hspace{0.0em}
            \includegraphics[height=1.0cm]{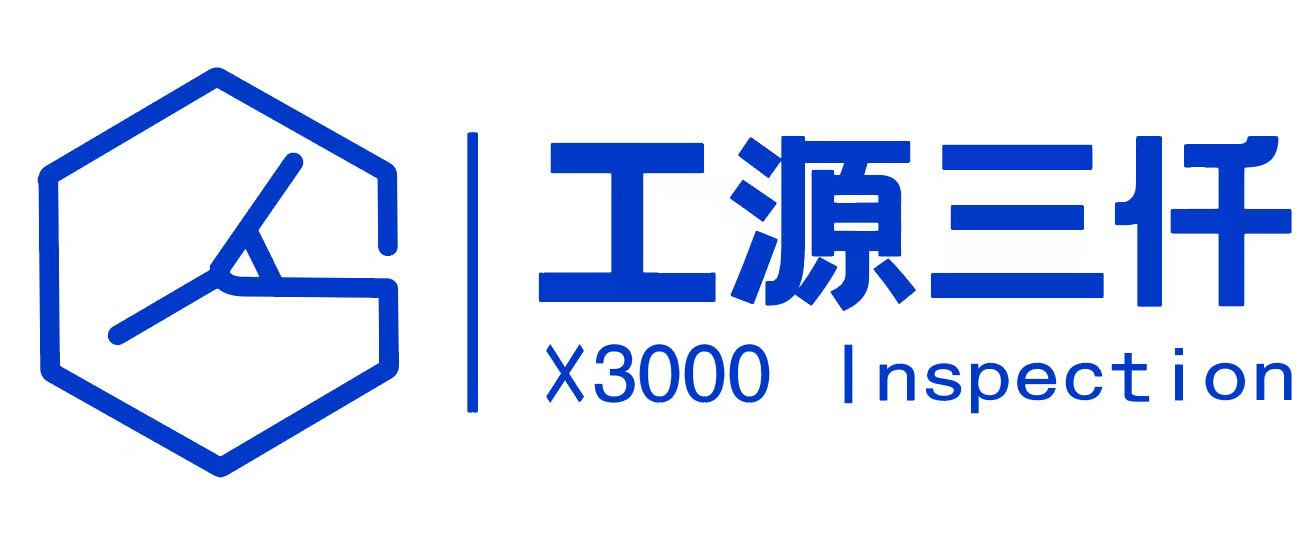}\hspace{0.1em}
		\end{minipage}
		\\[0.4em]
		\textcolor{academicblue}{\rule{\textwidth}{0.6pt}}%
	\end{minipage}%
}

\showlogointitle
\title{ConceptSeg-R1: Segment Any Concept via Meta-Reinforcement Learning}
\shorttitle{ConceptSeg-R1}

\makeatletter

\newcommand{\manualtitlecontent}{%
	\begin{center}
		{\MA@author@font
			Yuan Zhao$^{1,2\textbf{*}}$,
			Youwei Pang$^{3\textbf{*}}$,
			Jiaming Zuo$^{2}$,
			Wei Ji$^{4}$,
			\par\vspace{0.4em}
			Kailai Zhou$^{3}$,
			Bin Fan$^{5}$,
			Yunkang Cao$^{6}$,
			Lihe Zhang$^{1\boldsymbol{\ddag}}$,
			\par\vspace{0.4em}
			Xiaofeng Liu$^{4}$,
			Huchuan Lu$^{1}$,
			Weisi Lin$^{3}$,
			Dacheng Tao$^{3}$,
			Xiaoqi Zhao$^{3\boldsymbol{\ddag}}$
			\par}
		
		\vspace{\MA@author@affiliation@skip}
		
		{\MA@affiliation@font
			$^{1}$Dalian University of Technology, China\par
			$^{2}$X3000 Inspection Co., Ltd\par
			$^{3}$Nanyang Technological University, Singapore\par
			$^{4}$Yale University, USA\par
			$^{5}$Northwestern Polytechnical University, China\par
			$^{6}$Hunan University, China\par}
		
		\vspace{\MA@affiliation@email@skip}
		
		{\MA@email@font
			\textsuperscript{*}equal contribution.
			\textsuperscript{$\boldsymbol{\ddag}$}corresponding authors.
			\par}
	\end{center}
	
	\vspace{\MA@author@abstract@preskip}
	{\centering\textcolor{abstractborder}{\rule{0.3\textwidth}{0.4pt}}\par}
	\vspace{\MA@author@abstract@postskip}
	
	\noindent
Recent progress in promptable segmentation has shifted visual perception from object-level localization toward concept-level understanding. However, the notion of a concept remains under-specified, making it unclear whether current methods truly generalize beyond category recognition. 
In this work, we formalize generalized concept segmentation through a three-level taxonomy consisting of context-independent (CI), context-dependent (CD), and context-reasoning (CR) concepts, which reveals 
a clear capability gap across increasing levels of cognitive complexity. 
To address this challenge, we propose ConceptSeg-R1, a unified framework that reformulates concept segmentation as rule-induced concept grounding. 
At the core of our method is Meta-GRPO, a meta-reinforcement learning mechanism that learns transferable task rules from visual demonstrations and verifies them through proxy reasoning. The inferred reasoning states are then translated into segmentation-ready concept prompts via a lightweight concept translation module, enabling deductive application to target images.
A shortcut routing strategy further preserves the native efficiency of  segmentation models on simple cases. 
To systematically evaluate generalized concept segmentation, we conduct extensive experiments across diverse CI, CD, and CR concept segmentation benchmarks spanning natural, industrial, medical and reasoning-intensive domains. 
Without bells and whistles, ConceptSeg-R1 achieves strong performance across the full concept hierarchy while maintaining the native capability of promptable segmentation backbones. 
As an initial step toward segmenting any concept, we hope ConceptSeg-R1 can serve as a practical baseline for advancing segmentation from object-level prediction toward concept-level understanding.
	
	\par\vspace{0.4em}
	
	\begin{tabular}{@{}l@{}}
		\textbf{Project Page:} \textbf{\url{https://ntu-ai4x.github.io/ConceptSeg-R1}}\\
		\textbf{Source Code:} \textbf{\url{https://github.com/NTU-AI4X/ConceptSeg-R1}}
	\end{tabular}
}

\renewcommand{\@abstractboxcontent}{\manualtitlecontent}

\makeatother

\begin{document} 
	\maketitle

    \section{Introduction}
    \label{sec:intro}
Visual segmentation aims to localize regions that satisfy a user-specified semantic requirement. Over the past decade, the field has evolved from closed-set pixel classification with task-specific supervision~\cite{FCN,DeepLab,SegFormer,Mask2Former} toward open-world, prompt-driven perception.
The Segment Anything Model (SAM)~\cite{kirillov2023segment} marked a major milestone by unifying segmentation through geometric prompts such as points and boxes, enabling strong zero-shot transfer across diverse domains. More recently, SAM 3~\cite{carion2025sam} extended this paradigm to promptable concept segmentation, where targets can be specified using open-vocabulary concept prompts rather than fixed labels. This progress suggests a promising path toward more general segmentation systems.
However, a central question remains unresolved: \textit{what exactly is a concept in concept segmentation?} Existing works often equate concepts with text-addressable object categories, yet many real-world targets are defined not only by category identity, but also by contextual relations or reasoning requirements. Without a principled concept taxonomy, it remains difficult to determine whether a model truly generalizes beyond category recognition toward broader concept-level understanding.

\begin{figure}[t]
    \centering
    \includegraphics[width=1\linewidth]{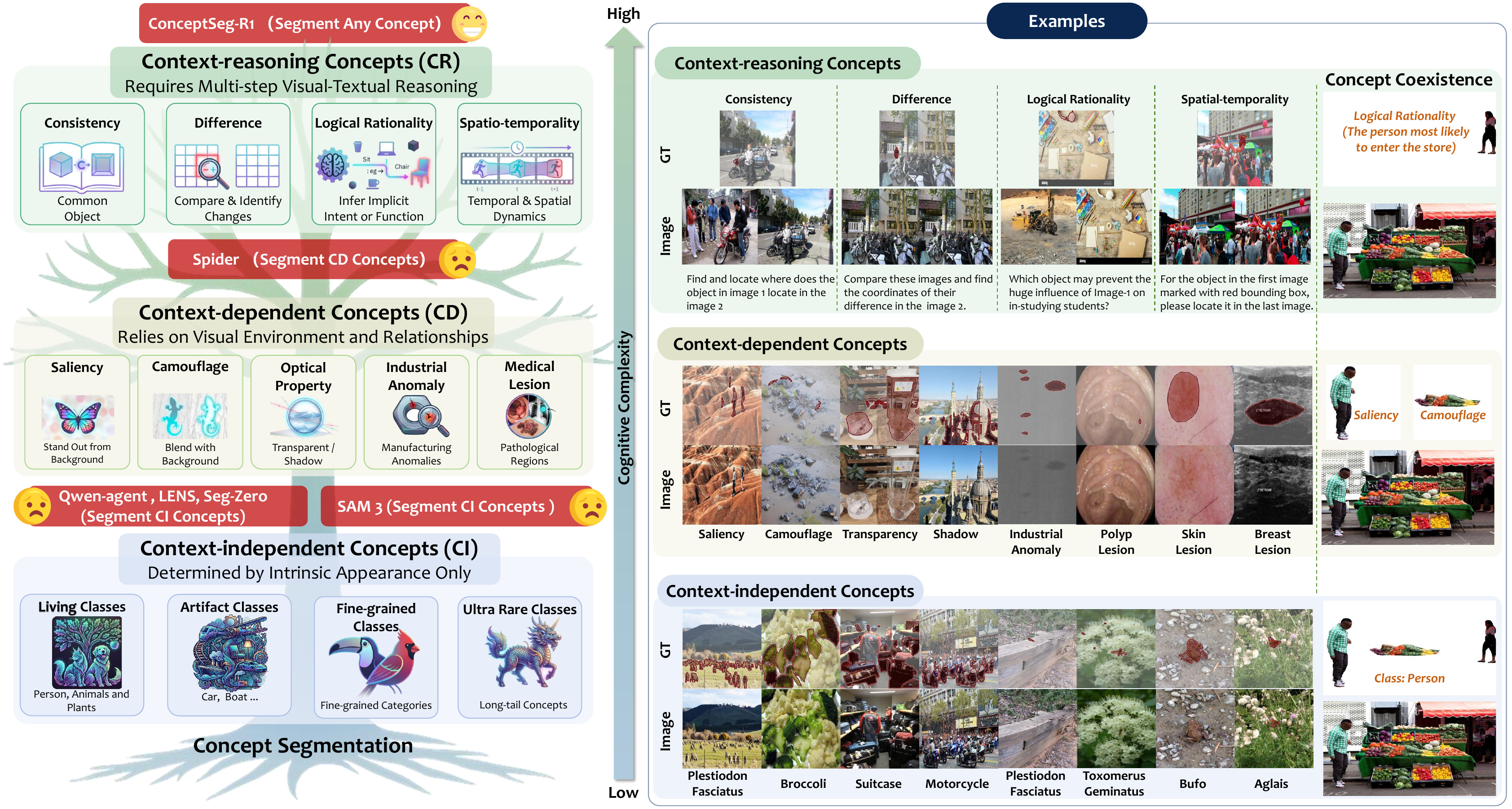}
    \caption{Concept segmentation tree  with representative CI, CD, and CR concepts arranged by cognitive complexity. 
        Existing promptable segmentation models mainly handle CI concepts, while context-oriented and reasoning-enhanced methods partially cover CD or CR concepts. 
        ConceptSeg-R1 targets the full concept hierarchy toward segmenting any concept. See \cref{appendix:Concept_Definition} for details.
    }
    \label{fig:ConceptSeg_Tree}
\end{figure}

Concept categorization has long been studied in philosophy and cognitive science, where concepts are  commonly distinguished as context-independent or context-dependent~\cite{CICD_1,CICD_2,CICD_3}. Similar distinctions have recently emerged in visual understanding and segmentation research~\cite{zhao2024spider,SAM_Eva}. However, these studies are  developed for task-specific models or narrowly defined prompting settings. In this work, we revisit concept segmentation from the perspective of general-purpose segmentation foundation models and methods based on multi-modal large language models (MLLMs), where concept understanding must account for varying levels of contextual dependency and cognitive complexity.

As illustrated in \cref{fig:ConceptSeg_Tree}, we organize generalized concept segmentation into a three-level concept hierarchy. 
The first level is \textbf{Context-Independent Concepts (CI)}, whose identities are determined primarily by intrinsic appearance and semantic attributes. General living, artifact,  fine-grained and long-tail ultra rare classes belong to this regime, where targets can often be recognized without strong contextual contrast cues. Promptable and generalist segmentation models~\cite{SegGPT,kirillov2023segment,carion2025sam} are naturally strong in this setting.
The second level is \textbf{Context-Dependent Concepts (CD)}, where targets are defined through their relation to the environment, such as saliency, camouflage, transparency, shadows, industrial anomalies, or medical lesions. These concepts require perception of contextual contrast or domain-specific structure rather than object identity alone. This perspective is closely related to Spider~\cite{zhao2024spider}, although it mainly studies context-dependent concepts under visual prompting rather than a unified text-and-vision interactive framework.
The third level is \textbf{Context-Reasoning Concepts (CR)}, which require explicit reasoning over visual and textual evidence, often involving cross-image correspondence, temporal cues, or functional semantics. 
These CR concepts go beyond contextual perception and require reasoning over relationships or functional dependencies across observations. 
Recent MLLM-based frameworks~\cite{li2025sam3,zhang2026tarot,liu2025seg,zhu2026lens} begin to explore this regime by integrating visual reasoning into segmentation.
Nevertheless, a unified concept hierarchy connecting CI, CD, and CR for general-purpose segmentation foundation models remains missing.

\begin{figure}[t]
    \centering
    \includegraphics[width=1\linewidth]{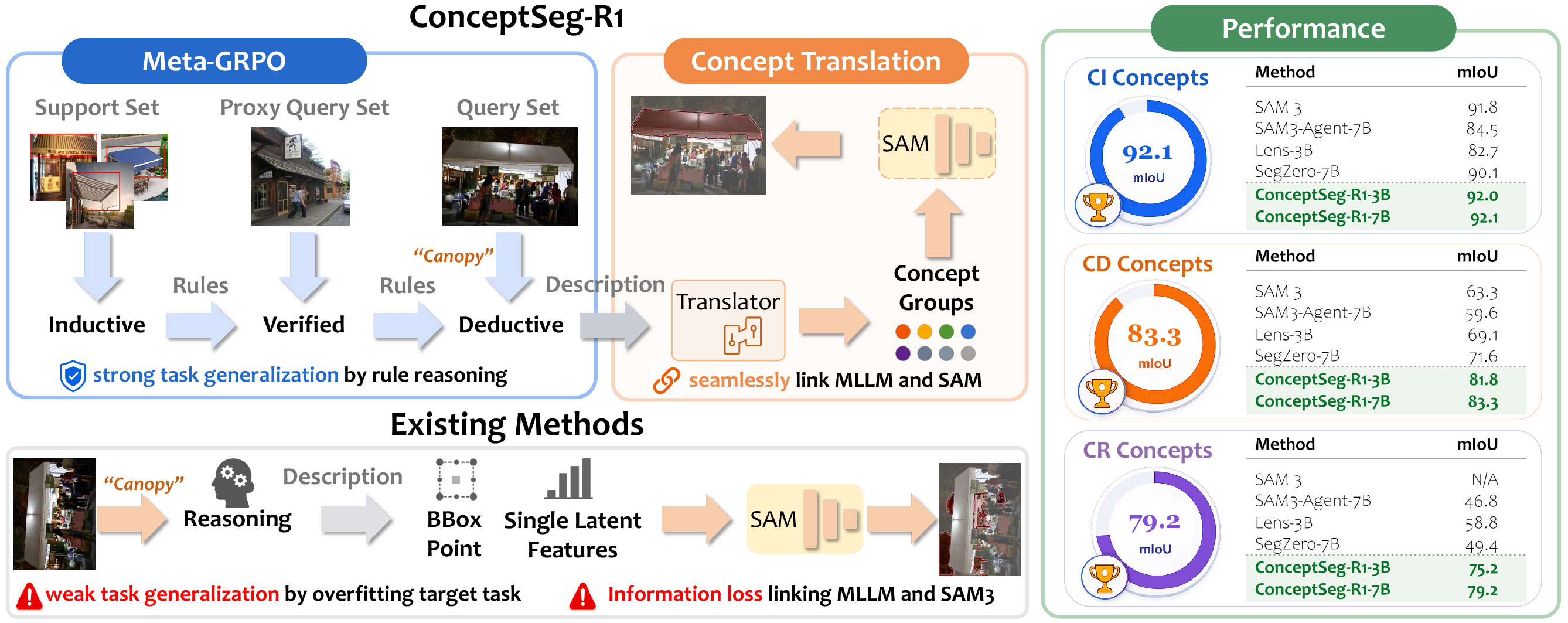}
    \caption{Workflow and performance comparison. ConceptSeg-R1 employs Meta-GRPO to induce  rules and deductive reasoning, which are converted into lossless prompts for SAM via concept translation, enabling robust full-spectrum concept segmentation. }
    \label{fig:teaser}
\end{figure}

Despite recent progress, existing methods still face two fundamental limitations, as illustrated by the ``existing methods'' pathway in \cref{fig:teaser}. 
First, many approaches follow a \emph{reasoning $\rightarrow$ description $\rightarrow$ interface $\rightarrow$ segmentation} pipeline, where rich semantics are compressed into low-bandwidth representations such as bounding boxes or short descriptions, causing a semantic bottleneck between reasoning and segmentation.
Second, most methods perform inference case by case rather than inducing transferable task rules from demonstrations. Instruction-tuning~\cite{LISA,li2025sam3,wei2025instructseg}, agent-based~\cite{carion2025sam,Segagent,MedSAM-Agent}, and reinforcement learning-based approaches~\cite{liu2025seg, zhu2026lens, you2025seg} improve reasoning consistency but still optimize instance-level trajectories instead of task-level induction. Consequently, they remain brittle under cross-task transfer, out-of-distribution shifts, and reasoning-intensive scenarios requiring concept inference from visual demonstrations.

In this work, we address these limitations with \textbf{ConceptSeg-R1}, a unified framework that reformulates generalized concept segmentation as a closed-loop process of \emph{task induction $\rightarrow$ rule verification $\rightarrow$ concept translation $\rightarrow$ promptable segmentation}, as illustrated in \cref{fig:teaser}. 
Instead of solving each example independently, ConceptSeg-R1 leverages reference images to induce transferable task rules and apply them to the query image. 
The first key component is \textbf{Meta-GRPO}, a meta-learning variant of group relative policy optimization that introduces a split-reference strategy with support examples and proxy queries. This mechanism enables the model to induce reusable rules from visual demonstrations, verify them on proxy queries, and apply them deductively to target images. 
The second key component is the \textbf{Concept Translation Module (CTM)}, which maps MLLM hidden states into implicit concept groups that can be injected into the segmentation prompt space, preserving rich reasoning signals for pixel-level execution. 
Finally, a \textbf{Shortcut Router} routes simple CI-style cases directly to SAM 3 while activating the full reasoning pipeline only when deeper contextual or logical inference is required. In this way, ConceptSeg-R1 unifies efficient promptable segmentation with context-aware concept grounding and rule-induced reasoning.

As shown in \cref{fig:teaser}, we evaluate ConceptSeg-R1  on  a total of $16$ benchmarks spanning CI, CD, and CR concepts.
By spanning the cognitive spectrum from fundamental recognition to visuospatial dependency modeling  and high-level visual-textual reasoning, this hierarchical evaluation validates the model’s effectiveness in bridging the gap between basic perception and complex task induction across diverse natural, industrial, and medical scenarios.
In addition, ConceptSeg-R1 achieves competitive results on widely used benchmarks such as Cityscapes~\cite{Cityscapes} and ReasonSeg~\cite{LISA}, further validating its generalization and reasoning capability. 
Taken together, these results suggest that generalized concept segmentation should be studied not merely as static category recognition, but as a unified problem of context-aware concept grounding, rule induction, and executable visual reasoning.

\section{Related Work}
\label{sec:related}

\textit{\textbf{Promptable Segmentation.}}
Recent advances have expanded visual segmentation from task-specific mask prediction to open-world interactive perception. 
SAM~\cite{kirillov2023segment} established a strong foundation by unifying diverse segmentation tasks through geometric prompts, such as points and boxes. 
Building on this paradigm, SAM 3~\cite{carion2025sam} introduced promptable concept segmentation, enabling segmentation with noun-phrase concepts and open-vocabulary semantic prompts. 
Recent methods further extend this capability to complex instructions: SAM3-I~\cite{li2025sam3} improves instruction following for SAM 3, while agent-based or reasoning-enhanced frameworks, such as AgentRVOS~\cite{jin2026agentrvos} and CoT-Seg~\cite{kao2026cot}, incorporate MLLMs and Chain-of-Thought reasoning for temporally or semantically complex queries. 
Despite these advances, most existing systems still follow a sequential reasoning-to-prompt pipeline, where rich intermediate reasoning is compressed into boxes, points, short descriptions, or single latent representations. 
This semantic bottleneck limits their ability to execute fine-grained reasoning at the mask level, especially for context-dependent and reasoning-intensive concepts. 

\textit{\textbf{MLLMs with Reinforcement Learning.}}
Multi-modal large language models (MLLMs) have recently been introduced into segmentation to enhance reasoning and instruction-following ability. 
Early reasoning segmentation frameworks, such as LISA~\cite{LISA} and GLaMM~\cite{GLaMM}, mainly rely on supervised fine-tuning to connect MLLM hidden states with mask decoders. 
Although effective for language-mask alignment, supervised training alone often struggles with out-of-distribution concepts, ambiguous instructions, and complex visual-textual reasoning. 
To improve reasoning consistency and spatial grounding, recent works have explored reinforcement learning for segmentation. 
Seg-Zero~\cite{liu2025seg} introduces GRPO-based optimization for reasoning-chain guided segmentation, while LENS~\cite{zhu2026lens} studies unified reinforced reasoning for segmentation. 
Subsequent methods, such as Seg-R1~\cite{you2025seg}, further incorporate dense spatial or mask-level feedback to improve grounding quality. 
These works demonstrate the potential of reinforcement learning for reasoning segmentation, but they primarily optimize instance-level reasoning trajectories, where each query is solved independently, leaving task-level transfer and rule induction insufficiently explored.

\textit{\textbf{Meta-Learning and Rule Induction.}}
Learning to adapt from limited examples has long been studied as a core objective of meta-learning. 
Classical methods such as MAML~\cite{finn2017model} and Reptile~\cite{nichol2018first} optimize model initialization for fast adaptation, while recent LLM studies connect this ability with in-context learning, where models infer task patterns from demonstrations without parameter updates. 
Works such as MetaICL~\cite{min2022metaicl} and MAML-en-LLM~\cite{sinha2024maml} improve few-shot generalization in language models. 
However, existing studies mainly focus on text-centric tasks or passive demonstration following, whereas generalized concept segmentation requires models to infer reusable visual rules from support examples, verify them under proxy contexts, and apply them to unseen target images. 
This paper brings this perspective into MLLM-based segmentation by formulating concept segmentation as rule-induced concept grounding, where visual demonstrations, proxy verification, and reinforced optimization jointly improve generalization across CI, CD, and CR concepts.

\section{ConceptSeg-R1 } 
\label{sec:method}

\subsection{Overall architecture}
\label{sec:architecture}

\begin{figure*}[t]
  \centering
   \includegraphics[width=\linewidth]{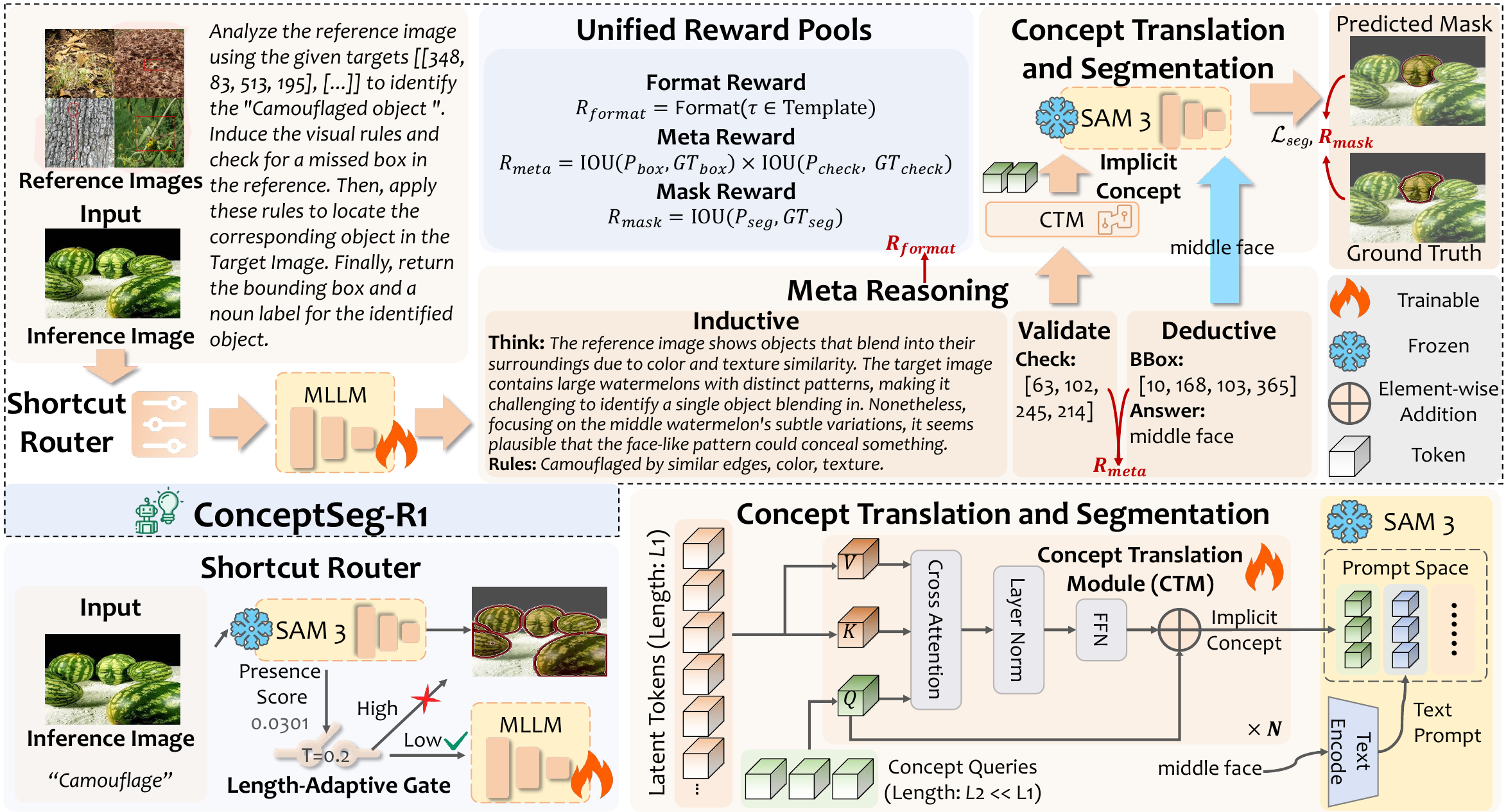}
   \caption{Overview of the ConceptSeg-R1 architecture. (\cref{sec:architecture})}
   \label{fig:Architecture}
\end{figure*}

We propose ConceptSeg-R1, a unified framework for generalized concept segmentation. 
The architecture follows a four-step paradigm:
\emph{task induction $\rightarrow$ rule verification $\rightarrow$ concept translation $\rightarrow$ promptable segmentation}.
As illustrated in \cref{fig:Architecture}, ConceptSeg-R1 utilizes an MLLM as the reasoning engine and SAM 3 as the segmentation backbone. %
The framework takes as input a mosaic of reference images (partitioned into a Support Set and a Proxy Query Set), a target inference image as the actual query, and a textual prompt.
A Shortcut Router adaptively determines whether full reasoning is needed: for high-confidence cases, SAM 3 is directly invoked via a fast path, bypassing meta-reasoning.
Otherwise, the model induces task-level rules from the Support Set, verifies them on the Proxy Query Set, and applies them to the target image.
Finally, the Concept Translation Module (CTM) maps these linguistic deductions into multi-dimensional implicit concept groups, serving as high-fidelity prompts for SAM 3 to generate precise pixel masks.

\subsection{Meta-GRPO}
\label{sec:meta_grpo}

We introduce Meta-GRPO, a meta-learning variant of GRPO~\cite{shao2024deepseekmath}, to improve in-context learning and out-of-distribution (OOD) generalization for generalized concept segmentation.
Unlike conventional segmentation methods that solve each image independently, Meta-GRPO internalizes meta-learning within a single visual context window through a Split-Reference Strategy. 
Given visual demonstrations, the model is encouraged to induce transferable task rules, verify them under proxy contexts, and deductively apply them to the target query image $\mathcal{I}$.
This formulation connects reinforcement learning with in-context meta-learning, enabling the model to move beyond instance-level reasoning toward rule-based concept grounding.

\textit{\textbf{Split-Reference Strategy.}}
To leverage multi-image context while maintaining computational efficiency, we organize the visual inputs into a mosaic-based reference structure. 
Given a target image $\mathcal{I}$, which serves as the actual query, and a textual prompt $\mathcal{T}$, we construct a reference set $\mathcal{R} = \{\mathcal{S}, \mathcal{Q}\}$ as a $K \times K$ mosaic image to reduce memory overhead. 
Within the mosaic, $\mathcal{S}$ denotes the Inductive Support Set, from which the model infers task-level rules, while $\mathcal{Q}$ denotes the Proxy Query Set, which is used for intermediate rule verification. 
The MLLM first performs induction over $\mathcal{S}$, then localizes the corresponding target in $\mathcal{Q}$ to validate the induced rules, and finally applies the verified rules to $\mathcal{I}$ by predicting object box coordinates and a descriptive linguistic answer.

\textit{\textbf{Meta Reward.}}
To encourage rule-based reasoning rather than shortcut memorization or reward hacking~\cite{amodei2016concrete}, we design a meta reward that jointly evaluates proxy-level verification and target localization quality.
The model samples trajectories $\tau$ from the concatenated visual-textual context:
\begin{equation}
\tau \sim \pi_{\theta}(\cdot \mid [\mathcal{S}, \mathcal{Q}, \mathcal{I}, \mathcal{T}])
\end{equation}
The meta reward is defined as
\begin{equation}
R_{{meta}} =
\text{IoU}(P_{box}, GT_{box})
\times
\text{IoU}(P_{check}, GT_{check})
\end{equation}
where $P_{box}$ and $GT_{box}$ denote the predicted and ground-truth boxes for $\mathcal{I}$, and $P_{check}$ and $GT_{check}$ denote the boxes for $\mathcal{Q}$. 
By coupling proxy deduction with target segmentation, Meta-GRPO rewards trajectories that rely on transferable, context-aware task rules rather than direct memorization of individual samples.

\subsection{Concept Translation}
\label{sec:concept_translation}

To eliminate the information bottleneck between high-level reasoning and segmentation execution, we develop a lightweight Concept Translation Module (CTM) with near-linear complexity. 
CTM converts the rich Chain-of-Thought (CoT) hidden states from the MLLM into multi-dimensional implicit concept groups that can be injected into the latent prompt space of SAM 3, enabling richer reasoning-to-mask transfer.

Specifically, given the MLLM output tokens $\mathbf{H} \in \mathbb{R}^{L_1 \times C}$, where $L_1$ is the token length, we initialize a compact set of learnable concept queries $\mathbf{C} \in \mathbb{R}^{L_2 \times C}$ with a much smaller length $L_2 \ll L_1$.
The concept queries attend to the reasoning tokens through cross-attention:
\begin{equation}
\mathbf{Z} = \mathrm{CrossAttn}(\mathbf{C}, \mathbf{H}, \mathbf{H})
\label{equ:crossattn_in_ctm}
\end{equation}
where $\mathbf{Z} \in \mathbb{R}^{L_2 \times C}$ denotes the extracted implicit concept groups.
These groups summarize multifaceted semantic attributes from the reasoning trajectory, such as appearance, spatial relations, relative scale, and functional properties.
Meanwhile, the final short linguistic description  is encoded by the SAM 3 text encoder to obtain explicit prompt embeddings $\mathbf{E}_{text}$. 
The implicit concept groups are then prepended to the explicit embeddings in the SAM 3 latent prompt space:
\begin{equation}
\mathbf{P} = [\mathbf{Z}; \mathbf{E}_{text}]
\end{equation}
where $\mathbf{P}$ serves as the final prompt representation for mask prediction.
Since CTM uses a small number of concept queries rather than the full MLLM token sequence, its cross-attention cost is $O(L_1L_2C)$ and becomes near-linear in $L_1$ when $L_2$ is fixed and much smaller than $L_1$.

Compared with conventional sequential pipelines that suffer from semantic disconnect, CTM translates complex reasoning states into segmentation-compatible visual guidance through a unified trainable interface. 
This replaces discrete tool-calling with continuous concept prompts, better preserving reasoning information while balancing segmentation accuracy and inference efficiency.

\subsection{Training and Inference}

We adopt a progressive training and adaptive inference strategy to balance reasoning capacity and computational efficiency.
Training consists of two stages: CTM warm-up with supervised segmentation loss, followed by Meta-GRPO optimization with unified rewards.
During inference, a lightweight Shortcut Router built upon the native confidence mechanism of SAM 3 decides whether to accept the direct SAM 3 prediction or activate the full ConceptSeg-R1 reasoning pipeline.

\textit{\textbf{Training Stage I: Supervised Fine-Tuning.}}
In the first stage, both the MLLM and SAM 3 are frozen, and only the CTM parameters are optimized.
This stage aligns MLLM reasoning features with the latent prompt space of SAM 3, allowing implicit concept groups to serve as effective segmentation prompts.
The optimization objective is the standard segmentation loss:
\begin{equation}
    \mathcal{L}_{{seg}}
    = \mathcal{L}_{{dice}}(P_{{seg}}, GT_{{seg}})
    + \mathcal{L}_{{focal}}(P_{{seg}}, GT_{{seg}})
\end{equation}
where $P_{{seg}}$ and $GT_{{seg}}$ denote the predicted and ground-truth masks, respectively.

\textit{\textbf{Training Stage II: Cognitive Reinforcement Learning.}}
In the second stage, the reasoning policy $\pi_{\theta}$ is optimized with Meta-GRPO under a structured CoT template consisting of
\texttt{<rule>}, \texttt{<check>}, \texttt{<think>}, \texttt{<bbox>}, and \texttt{<answer>} tags.
The unified reward pool combines format correctness, mask quality, and meta-level rule verification:
\begin{equation}
    R_{uni} = R_{format} + R_{mask} + R_{meta}
\end{equation}
The format and mask rewards are defined as:
\begin{equation}
    R_{format} = \mathbf{1} \left[ t_{\tau} \models \mathcal{T}_{CoT} \right]
    \quad
    R_{mask} = \text{IoU}(P_{seg}, GT_{seg})
\end{equation}
where $t_{\tau}$ denotes the generated reasoning response of trajectory $\tau$, and $t_{\tau} \models \mathcal{T}_{\text{CoT}}$ indicates that the response satisfies the required CoT template.
Here, $R_{meta}$ is the proxy-target reward defined in \cref{sec:meta_grpo}, which couples proxy rule verification with target localization quality.
The final training objective combines Meta-GRPO optimization with the segmentation loss:
\begin{equation}
    \mathcal{L}_{train} = \mathcal{L}_{GRPO}(R_{uni}, D_{KL}) + \mathcal{L}_{seg}
\end{equation}
where KL divergence term $D_{KL}$ acts as a standard regularization constraint, preventing the current policy from deviating excessively from the reference policy.
This ensures linguistic stability and prevents catastrophic forgetting by anchoring the reasoning trajectories within a plausible distribution.

\textit{\textbf{Adaptive Inference with Shortcut Router.}}
At inference time, the Shortcut Router first uses the native SAM 3 prediction branch to obtain a presence score $s$, which reflects the confidence that the queried concept can be directly localized without additional reasoning. 
The decision threshold $T$ is adjusted according to instruction complexity:
\begin{equation}
    T = \mathrm{CLIP}(0.1 \times 2^{\ell(\mathcal{T})-1}, \min=0, \max=1)
\end{equation}
where $\ell(\mathcal{T})$ denotes the normalized word count of the instruction $\mathcal{T}$, and $\text{CLIP}(\cdot)$ denotes the clamping function used to constrain the output within a predefined range.
A smaller $T$ is produced for short atomic instructions, making SAM 3 predictions easier to accept through the shortcut path when $s \geq T$. 
In contrast, longer queries yield a larger $T$, imposing a stricter acceptance criterion and activating the full ConceptSeg-R1 pipeline when native SAM 3 confidence is insufficient. 
This adaptive mechanism allows simple CI-style queries to be handled efficiently while reserving full reasoning capacity for complex CD and CR concepts.

\section{Experiments}
\label{sec:experiment}

\subsection{Settings}

\textit{\textbf{Datasets.}}
We evaluate ConceptSeg-R1 on a diverse set of concept segmentation benchmarks spanning context-independent (CI), context-dependent (CD), and context-reasoning (CR) concepts. 
These benchmarks cover natural, industrial, medical, and reasoning-intensive scenarios, enabling a comprehensive assessment of out-of-distribution generalization across different levels of cognitive complexity. 
In addition, we include representative generic segmentation and reasoning-oriented benchmarks, such as Cityscapes~\cite{Cityscapes} and ReasonSeg~\cite{LISA}, to verify whether ConceptSeg-R1 preserves the generic segmentation capability of SAM 3 and generalizes to external reasoning-grounded settings.
Complete dataset descriptions and evaluation protocols are provided in \cref{appendix:Benchmark}.

\textit{\textbf{Implementation Details and Metrics.}}
We employ the Qwen2.5-VL series~\cite{bai2025qwen25vltechnicalreport} as the reasoning backbone and SAM 3 as the segmentation head, using both 3B and 7B variants in our experiments.  Training is conducted on 8 NVIDIA H800 GPUs utilizing the DeepSpeed engine to optimize memory and training efficiency.
Following previous works~\cite{carion2025sam,zhu2026lens,liu2025seg,zhao2024spider}, 
we report four widely adopted segmentation metrics: weighted F-measure ({\tt $F_{\beta}^{\omega}$}), mean Intersection-over-Union ({\tt mIoU}), generalized IoU ({\tt gIoU})
and cumulative IoU ({\tt cIoU}). 
All results are reported in percentage (\%). 
More details and metrics are provided in \cref{appendix:Implementation} and \cref{appendix:detailed_results}.

\subsection{Comparison}

\begin{table*}[t]
    \centering
     \begin{minipage}{\linewidth}
     
    \caption{Quantitative comparison on diverse CI, CD, and CR concept segmentation benchmarks. 
Detailed results for diverse classes in CI concepts and more evaluation metrics are provided in \cref{appendix:detailed_results}.}
    \begin{adjustbox}{width=\linewidth}

\begin{tabular}{r *{22}{c}}
& \multicolumn{2}{c}{\textit{CI Concepts}}
& \multicolumn{10}{c}{\textit{CD Concepts}}
& \multicolumn{8}{c}{\textit{CR Concepts}}
& \multicolumn{2}{c}{\multirow{2}{*}[-5ex]{\textbf{Mean}}} \\
\cmidrule(lr){2-3}\cmidrule(lr){4-13}\cmidrule(lr){14-21}
& \multicolumn{2}{c}{\makecell{\scriptsize Diverse\\\scriptsize Classes}}
& \multicolumn{2}{c}{\makecell{\scriptsize Optical \\\scriptsize Property}}
& \multicolumn{2}{c}{\makecell{\scriptsize Camouflage}}
& \multicolumn{2}{c}{\makecell{\scriptsize Saliency}}
& \multicolumn{2}{c}{\makecell{\scriptsize Industrial\\\scriptsize Anomaly}}
& \multicolumn{2}{c}{\makecell{\scriptsize Medical\\\scriptsize Lesion}}
& \multicolumn{2}{c}{\makecell{\scriptsize Consistency}}
& \multicolumn{2}{c}{\makecell{\scriptsize Difference}}
& \multicolumn{2}{c}{\makecell{\scriptsize Logical\\\scriptsize Rationality}}
& \multicolumn{2}{c}{\makecell{\scriptsize Spatio-temporality}}
& \multicolumn{2}{c}{} \\
\cmidrule(lr){2-3}\cmidrule(lr){4-5}\cmidrule(lr){6-7}\cmidrule(lr){8-9}\cmidrule(lr){10-11}\cmidrule(lr){12-13}\cmidrule(lr){14-15}\cmidrule(lr){16-17}\cmidrule(lr){18-19}\cmidrule(lr){20-21}\cmidrule(lr){22-23}
{\textbf{Method}}
& {\tt\small $F_{\beta}^{\omega} \uparrow$} & {\tt\small mIoU $\uparrow$}  
& {\tt\small $F_{\beta}^{\omega} \uparrow$} & {\tt\small mIoU $\uparrow$}
& {\tt\small $F_{\beta}^{\omega} \uparrow$} & {\tt\small mIoU $\uparrow$}
& {\tt\small $F_{\beta}^{\omega} \uparrow$} & {\tt\small mIoU $\uparrow$}
& {\tt\small $F_{\beta}^{\omega} \uparrow$} & {\tt\small mIoU $\uparrow$}
& {\tt\small $F_{\beta}^{\omega} \uparrow$} & {\tt\small mIoU $\uparrow$}
& {\tt\small $F_{\beta}^{\omega} \uparrow$} & {\tt\small mIoU $\uparrow$}
& {\tt\small $F_{\beta}^{\omega} \uparrow$} & {\tt\small mIoU $\uparrow$}
& {\tt\small $F_{\beta}^{\omega} \uparrow$} & {\tt\small mIoU $\uparrow$}
& {\tt\small $F_{\beta}^{\omega} \uparrow$} & {\tt\small mIoU $\uparrow$}
& {\tt\small $F_{\beta}^{\omega} \uparrow$} & {\tt\small mIoU $\uparrow$}\\
\toprule

SAM 3~\cite{carion2025sam}
  & \underline{89.5} & {91.8}   
  & \underline{85.6} & \underline{86.4}  & 51.3 & 61.4
  & 38.1 & 59.0  & 51.5 & 66.2
  & 36.3 & 48.9
  & -- & --  & -- & --  & -- & --  & -- & --
  & -- & -- \\

\rowcolor{mygray}
\multicolumn{23}{c}{\textbf{\textit{MLLM + SAM}}} \\
\midrule
SAM3-Agent-3B~\cite{carion2025sam}
  & 55.1 & 72.3   
  & 47.5 & 62.0  & 45.4 & 64.8
  & 59.5 & 72.1  & 20.6 & 49.1
  & 17.1 & 41.4
  & 20.3 & 43.2  & 3.1  & 30.9  & 20.3 & 46.9  & 10.2 & 40.5
  & \meanvalue{29.9} & \meanvalue{52.3} \\
SAM3-Agent-7B~\cite{carion2025sam}
  & 76.8 & 84.5   
  & {68.3} & {74.3}  & 58.9 & 71.7
  & 74.4 & 80.2  & 31.5 & 49.0
  & 25.1 & 42.5
  & 26.7 & 46.8  & 26.7 & 53.1  & 29.3 & 53.3  & 13.5 & 37.1
  & \meanvalue{43.1} & \meanvalue{59.3} \\
LENS-3B~\cite{zhu2026lens}
  & 74.4 & 82.7   
  & 56.0 & 65.4  & 58.1 & 73.1
  & 76.9 & 81.1  & 51.3 & 60.2
  & 56.0 & 69.1
  & 26.7 & 55.2  & 25.6 & 56.8  & 19.5 & 55.1  & 43.7 & 68.2
  & \meanvalue{48.8} & \meanvalue{66.7} \\
Seg-Zero-7B~\cite{liu2025seg}
  & {86.8} & 90.1   
  & 67.1 & 73.1  & 75.0 & 81.1
  & 72.8 & 78.6  & 51.4 & 57.1
  & 60.2 & 69.8
  & 16.1 & 49.6  & 5.5  & 49.2  & 9.8  & 49.3  & 7.2  & 49.3
  & \meanvalue{45.2} & \meanvalue{64.7} \\
\rowcolor{Light}
{ConceptSeg-R1-3B}
  & \textbf{89.9} & \underline{92.0}
  & \textbf{85.7} & \textbf{86.5}
  & \underline{83.7} & \underline{87.7}
  & \underline{89.0} & \underline{90.5}
  & \underline{61.5} & \underline{71.0}
  & \underline{69.0} & \underline{77.4}
  & \underline{63.9} & \underline{77.0}
  & \underline{52.7} & \underline{71.8}
  & \underline{46.6} & \underline{68.5}
  & \underline{64.7} & \underline{79.1}
  & \underline{\meanvalue{70.7}} & \underline{\meanvalue{80.1}} \\
\rowcolor{Light}
{ConceptSeg-R1-7B}
  & \textbf{89.9} & \textbf{92.1} 
  & \textbf{85.7} & \textbf{86.5} 
  & \textbf{84.8} & \textbf{88.3} 
  & \textbf{92.7} & \textbf{93.5} 
  & \textbf{66.7} & \textbf{74.1} 
  & \textbf{72.3} & \textbf{79.3} 
  & \textbf{70.1} & \textbf{81.0} 
  & \textbf{57.0} & \textbf{75.0}  
  & \textbf{60.2} & \textbf{76.7} 
  & \textbf{69.8} & \textbf{81.8}
  & \textbf{\meanvalue{74.9}} & \textbf{\meanvalue{82.8}} \\
  \bottomrule
\end{tabular}
    \end{adjustbox}
    \label{tab:quantitative_comparison}
    \end{minipage}
\end{table*}

\textit{\textbf{Performance on CI, CD and CR Concepts.}}
As summarized in \cref{tab:quantitative_comparison}, ConceptSeg-R1 achieves consistently strong performance across the full CI–CD–CR spectrum. 
On CI concepts, SAM 3 first shows strong segmentation capability, even outperforming several MLLM-based models such as LENS and SegZero. However, directly pairing SAM 3 with an MLLM without careful design can lead to noticeable performance degradation, as evidenced by the lower results of SAM3-Agent-3B and SAM3-Agent-7B. In contrast, ConceptSeg-R1 preserves the strong segmentation capability of SAM 3 while achieving the best overall performance on diverse CI concepts.
On CD concepts, ConceptSeg-R1 shows clear advantages across diverse domains, including optical properties, camouflage, industrial anomaly, and medical lesions. Notably, ConceptSeg-R1-3B achieves 77.4 {\tt mIoU} on the medical lesion benchmark, outperforming the larger Seg-Zero-7B model.
On CR concepts, SAM 3 is not designed to handle scenarios that require reasoning across multiple observations, and therefore cannot directly perform such tasks. Among MLLM-based methods, ConceptSeg-R1 shows the most significant performance advantage on CR concepts, consistently achieving the best results on consistency, difference, logical rationality, and spatio-temporality reasoning concepts. 

\begin{table*}[t]
    \centering
    \begin{minipage}{\linewidth}
        \caption{Comparison of ConceptSeg-R1 and SAM 3 on Cityscapes under the zero-shot setting, evaluated using the {\tt mIoU} metric. Since official semantic segmentation scripts for SAM 3 are unavailable, results are reported under the same inference settings as ConceptSeg-R1 for fair comparison.}
        \label{tab:cityscapes}
        \begin{adjustbox}{width=\linewidth}
            \begin{tabular}{r*{19}{c}c}
    \textbf{Method}            & {Road}            & {Side.}           & {Build.}          & {Wall}            & {Fence}           & {Pole}            & {T.Light}         & {T.Sign}          & {Veget.}          & {Terra.}          & {Sky}             & {Person}          & {Rider}           & {Car}             & {Truck}            & {Bus}             & {Train}            & {Motor.}          & {Bicycle}         & \textbf{Mean}            \\
    \toprule
    SAM 3~\cite{carion2025sam} & 91.4                     & 82.2                     & 84.4                     & 19.1                     & 45.4                     & 61.1                     & 64.7                     & 67.2                     & 88.4                     & 5.7                      & 92.6                     & 76.2                     & 39.3                     & 87.6                     & 42.6                      & 54.3                     & 29.2                      & 46.6                     & 72.9                     & \meanvalue{60.6}                     \\
    ConceptSeg-R1-3B           & 97.9                     & 82.6                     & 83.9                     & 18.8                     & 47.3                     & 60.8                     & 60.2                     & 66.6                     & 87.3                     & 7.2                      & 93.7                     & 74.8                     & 39.4                     & 89.4                     & 52.8                      & 62.3                     & 42.6                      & 50.5                     & 72.1                     & \meanvalue{62.6}                     \\
    \midrule
    $\Delta$ Gains             & \textbf{\increase{+6.5}} & \textbf{\increase{+0.4}} & \textbf{\decrease{-0.5}} & \textbf{\decrease{-0.3}} & \textbf{\increase{+1.9}} & \textbf{\decrease{-0.3}} & \textbf{\decrease{-4.5}} & \textbf{\decrease{-0.6}} & \textbf{\decrease{-1.1}} & \textbf{\increase{+1.5}} & \textbf{\increase{+1.1}} & \textbf{\decrease{-1.4}} & \textbf{\increase{+0.1}} & \textbf{\increase{+1.8}} & \textbf{\increase{+10.2}} & \textbf{\increase{+8.0}} & \textbf{\increase{+13.4}} & \textbf{\increase{+3.9}} & \textbf{\decrease{-0.8}} & \textbf{\increase{+2.0}} \\
      \bottomrule
\end{tabular}

        \end{adjustbox}
    \end{minipage}
\end{table*}

\textit{\textbf{Performance on Cityscapes and ReasonSeg.}}
As shown in \cref{tab:cityscapes}, ConceptSeg-R1 maintains strong zero-shot segmentation performance on the Cityscapes benchmark, improving the overall {\tt mIoU} from 60.6 to 62.6 without task-specific training. 
In addition, \cref{tab:zeroshot_reasonseg} shows that ConceptSeg-R1 achieves state-of-the-art performance on the ReasonSeg benchmark. 
Notably, our method achieves both the highest {\tt gIoU} and {\tt cIoU} on ReasonSeg-Test without task-specific fine-tuning, demonstrating strong reasoning capability and robust generalization to external reasoning-oriented segmentation settings.

\begin{wraptable}[12]{r}{0.44\linewidth}
    \centering
    \vspace{-5mm}
    \caption{Zero-shot performance on ReasonSeg~\cite{LISA}.
    ``$\dagger$'' denotes models fine-tuned on the ReasonSeg training set.}
    \label{tab:zeroshot_reasonseg}
    \begin{adjustbox}{width=\linewidth} 
        \begin{tabular}{rccccc}


& & \multicolumn{2}{c}{\textbf{ReasonSeg-Val}} 

& \multicolumn{2}{c}{\textbf{ReasonSeg-Test}} \\

\cmidrule(lr){3-4} \cmidrule(lr){5-6}

\textbf{Method} & \textbf{Publication}

& {\tt\small gIoU $\uparrow$} & {\tt\small cIoU $\uparrow$} & {\tt\small gIoU $\uparrow$} & {\tt\small cIoU $\uparrow$} \\

\toprule

LISA-7B$\dagger$~\cite{LISA} & CVPR'24  & 52.9 & 54.0 & 55.6 & 56.9 \\

InstructSeg-3B$\dagger$~\cite{wei2025instructseg} & ICCV'25  & 61.9 & {65.2} & --  & --  \\

LENS-3B$\dagger$~\cite{zhu2026lens}  & AAAI'26  & 62.1 & {64.9} & 57.2 & {58.0} \\
\midrule 
SAM4MLLM-7B~\cite{SAM4MLLM} & ECCV'24  & 46.7 & 48.1 & -- & -- \\
Seg-Zero-3B~\cite{liu2025seg} & arXiv'25  & 58.2 & 53.1 & 56.1 & 48.6 \\

Seg-Zero-7B~\cite{liu2025seg} & arXiv'25 & 62.6 & \textbf{62.0} & 57.5 & 52.0 \\

SAM-R1-7B~\cite{huang2026samr1leveragingsamreward}  & NeurIPS'25  & \underline{64.0} & 55.8 & 60.2 & 54.3 \\

SAM3-Agent-7B~\cite{carion2025sam} & ICLR'26  & 62.2 & 49.1 & \textbf{63.0} & 53.5 \\

DPAD-7B~\cite{yang2026discriminative}  & CVPR'26  & 63.1 & \underline{61.2} & 57.7 & \underline{54.4} \\


\rowcolor{Light}

ConceptSeg-R1-3B & --& 62.8 & 54.0 &\underline{61.2} & 49.3\\

\rowcolor{Light}

ConceptSeg-R1-7B &-- &  \textbf{64.4}& 55.1 &\textbf{63.0} & \textbf{59.3}\\

\bottomrule

\end{tabular}
    \end{adjustbox}
    \vspace{-5em}
\end{wraptable}

\begin{figure*}[t!]
   \centering
   \includegraphics[width=\linewidth]{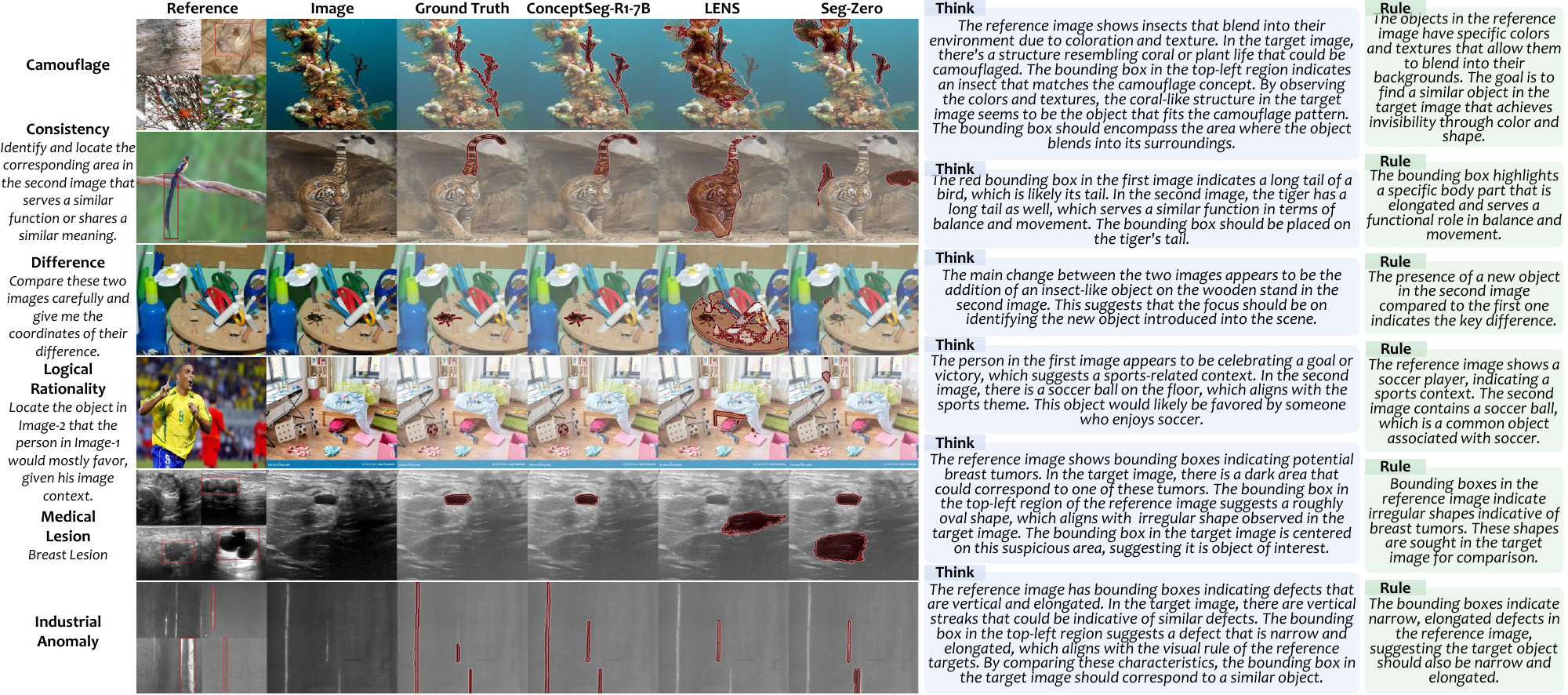}
   \caption{Qualitative comparison across different concepts. See \cref{appendix:Visualization} for more visualization results.}
   \label{fig:Qualitative_Comparison}
\end{figure*} 

\begin{figure*}[t]
   \centering
   \includegraphics[width=\linewidth]{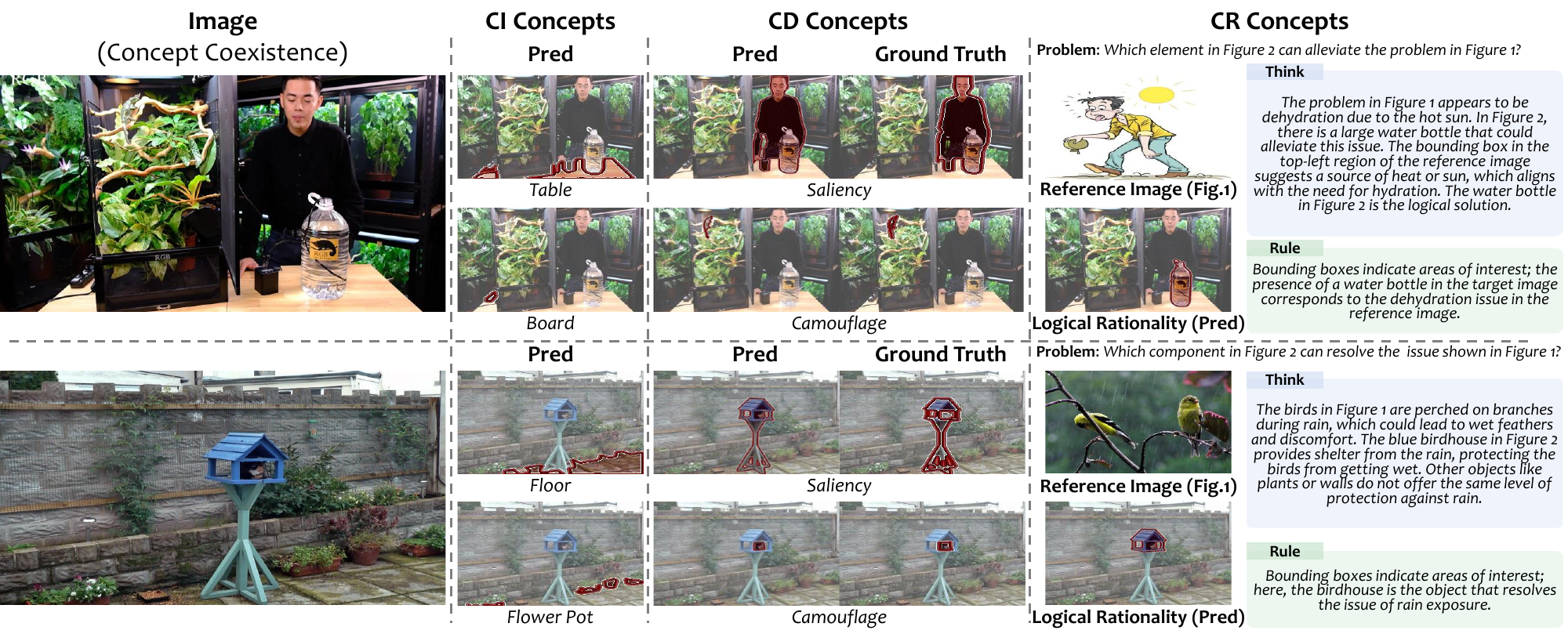}
   \caption{Visualization of concept coexistence results generated by ConceptSeg-R1.
   The same query image is segmented differently under varying reference instructions.}
   \label{fig:Concept_conexistence_visual}
\end{figure*}

\textit{\textbf{Qualitative Comparison.}}
As shown in \cref{fig:Qualitative_Comparison}, ConceptSeg-R1 shows superior visual grounding compared to LENS and Seg-Zero, highlighting three key advantages:
\textit{{\uppercase\expandafter{\romannumeral1}) Rule-based Generalization:}}
For camouflaged object and medical lesion concepts (see the 1$^{st}$ and 5$^{th}$ rows in Fig.~\ref{fig:Qualitative_Comparison}), ConceptSeg-R1 produces more precise boundaries while competing models often generate fragmented masks, indicating that Meta-GRPO effectively induces transferable task rules for domain-specific perception.
\textit{{\uppercase\expandafter{\romannumeral2}) Cross-image Reasoning:}} 
In consistency and difference reasoning tasks, ConceptSeg-R1 correctly identifies functional correspondences and inter-class differences across images, suggesting that the split-reference strategy enables stable cross-image reasoning.
\textit{{\uppercase\expandafter{\romannumeral3}) High-Fidelity Execution:}}  
In logical reasoning scenarios, the concept translation module  preserves rich reasoning states for segmentation, avoiding the semantic loss commonly observed in methods relying on coarse prompts. 
This lossless concept translation also empowers our model to excel in multi-object segmentation and demonstrates a superior capacity to rectify suboptimal ground-truth labels, such as missing targets or imprecise boundaries in industrial anomaly scenarios (see the 6$^{th}$ row in Fig.~\ref{fig:Qualitative_Comparison}).
To further demonstrate the prompt understanding capability of ConceptSeg-R1, \cref{fig:Concept_conexistence_visual} presents representative predictions under concept coexistence scenarios. 
We observe that ConceptSeg-R1 learns to switch segmentation targets accurately according to different prompts, without relying on a specific query image. 
In particular, for CR concepts, the model infers that dehydration requires water and that birds exposed to rain require shelter, selecting the water bottle and birdhouse accordingly, which demonstrates reasoning-driven target selection guided by reference understanding and problem interpretation.

\begin{table*}[t!]
    \centering
    \begin{minipage}{\linewidth}
        \caption{Ablation studies of ConceptSeg-R1-3B across the full CI-CD-CR concept spectrum.}
        \label{tab:overall_ablation}
        \begin{adjustbox}{width=\linewidth}
            
\begin{tabular}{l *{22}{c}}
& \multicolumn{2}{c}{\textit{CI Concepts}}
& \multicolumn{10}{c}{\textit{CD Concepts}}
& \multicolumn{8}{c}{\textit{CR Concepts}}
& \multicolumn{2}{c}{\multirow{2}{*}[-5ex]{\textbf{Mean}}}
\\
\cmidrule(lr){2-3}\cmidrule(lr){4-13}\cmidrule(lr){14-21}
& \multicolumn{2}{c}{\makecell{\scriptsize  Diverse \\\scriptsize Classes}}
& \multicolumn{2}{c}{\makecell{\scriptsize Optical  \\\scriptsize Property}}
& \multicolumn{2}{c}{\makecell{\scriptsize Camouflage}}
& \multicolumn{2}{c}{\makecell{\scriptsize Saliency}}
& \multicolumn{2}{c}{\makecell{\scriptsize Industrial\\\scriptsize Anomaly}}
& \multicolumn{2}{c}{\makecell{\scriptsize Medical\\\scriptsize Lesion}}
& \multicolumn{2}{c}{\makecell{\scriptsize Consistency}}
& \multicolumn{2}{c}{\makecell{\scriptsize Difference}}
& \multicolumn{2}{c}{\makecell{\scriptsize Logical\\\scriptsize Rationality}}
& \multicolumn{2}{c}{\makecell{\scriptsize Spatio-temporality}}
& \multicolumn{2}{c}{} 
\\
\cmidrule(lr){2-3}\cmidrule(lr){4-5}\cmidrule(lr){6-7}\cmidrule(lr){8-9}\cmidrule(lr){10-11}\cmidrule(lr){12-13}\cmidrule(lr){14-15}\cmidrule(lr){16-17}\cmidrule(lr){18-19}\cmidrule(lr){20-21}\cmidrule(lr){22-23}
{\textbf{Method}}
& {\tt\small $F_{\beta}^{\omega} \uparrow$} & {\tt\small mIoU $\uparrow$} 
& {\tt\small $F_{\beta}^{\omega} \uparrow$} & {\tt\small mIoU $\uparrow$} 
& {\tt\small $F_{\beta}^{\omega} \uparrow$} & {\tt\small mIoU $\uparrow$}
& {\tt\small $F_{\beta}^{\omega} \uparrow$} & {\tt\small mIoU $\uparrow$} 
& {\tt\small $F_{\beta}^{\omega} \uparrow$} & {\tt\small mIoU $\uparrow$}
& {\tt\small $F_{\beta}^{\omega} \uparrow$} & {\tt\small mIoU $\uparrow$}
& {\tt\small $F_{\beta}^{\omega} \uparrow$} & {\tt\small mIoU $\uparrow$} 
& {\tt\small $F_{\beta}^{\omega} \uparrow$} & {\tt\small mIoU $\uparrow$}
& {\tt\small $F_{\beta}^{\omega} \uparrow$} & {\tt\small mIoU $\uparrow$} 
& {\tt\small $F_{\beta}^{\omega} \uparrow$} & {\tt\small mIoU $\uparrow$}
& {\tt\small $F_{\beta}^{\omega} \uparrow$} & {\tt\small mIoU $\uparrow$}
\\
\toprule

\rowcolor{mygray}
\multicolumn{23}{c}{\textbf{\textit{(a) Architecture}}} \\
\midrule
 SAM 3 
  &  89.5  & 91.8  
  & {85.6} &  86.4   & 51.3 & 61.4
  & 38.1 & 59.0  & 51.5 & 66.2
  & 36.3 & 48.9
  & -- & --  & -- & --  & -- & --  & -- & --
  & -- & -- \\
+ MLLM (SFT)
& 89.2 & 91.4 & 85.8 & 86.5 & 80.6 & 85.6 & 87.2 & 88.4 & 55.9 & 61.9 & 57.3 & 67.5 & 28.3 & 56.8 & 15.9 & 53.1 & 18.3 & 53.4 & 21.7 & 56.8 & \meanvalue{54.0} & \meanvalue{70.1} \\
+ GRPO
& 89.4 & 91.6 & 85.8 & 86.5 & 76.4 & 82.3 & 86.4 & 87.8 & 55.5 & 62.4 & 54.5 & 65.0 & 25.2 & 54.2 & 21.4 & 53.5 & 32.2 & 60.0 & 30.3 & 61.4 & \meanvalue{55.7} & \meanvalue{70.5} \\
+ Concept Translations
& 89.5 & 91.8 & \textbf{85.8} & {86.5} & 82.0 & 86.5 & \textbf{90.7} & \textbf{91.8} & 56.5 & 63.7 & 61.8 & 70.2 & 58.5 & 73.4 & 31.4 & 54.9 & 45.6 & 66.5 & 57.6 & 75.5 & \meanvalue{65.9} & \meanvalue{75.8} \\
\rowcolor{Light}
+ Meta-GRPO
& \textbf{89.9} & \textbf{92.0} & 85.7 & \textbf{86.5} & \textbf{83.7} & \textbf{87.7} & 89.0 & 90.5 & \textbf{61.5} & \textbf{71.0} & \textbf{69.0} & \textbf{77.4} & \textbf{63.9} & \textbf{77.0} & \textbf{52.7} & \textbf{71.8} & \textbf{46.6} & \textbf{68.5} & \textbf{64.7} & \textbf{79.1} & \textbf{\meanvalue{70.7}} & \textbf{\meanvalue{80.1}} \\

\rowcolor{mygray}
\multicolumn{23}{c}{\textbf{\textit{(b) Meta-GRPO Component}}} \\
\midrule
w/o Meta Reward
& 89.5 & 91.8 & \textbf{85.8} & {86.5} & 82.4 & 86.6 & 89.2 & 90.4 & 60.0 & 66.6 & 68.1 & 76.9 & 59.6 & 74.1 & 51.7 & 71.1 & 41.8 & 66.8 & \textbf{65.4} & \textbf{79.5} & \meanvalue{69.3} & \meanvalue{79.0} \\
w/o Meta Reasoning
& {89.9} & {92.0} & 85.7 & {86.5} & 83.6 & 87.6 & \textbf{90.3} & \textbf{91.3} & 60.4 & 70.4 & 66.1 & 76.2 & 61.3 & 75.5 & 48.5 & 68.8 & 36.2 & 63.3 & 63.5 & 78.5 & \meanvalue{68.5} & \meanvalue{79.0} \\
\rowcolor{Light}
ConceptSeg-R1
& \textbf{89.9} & \textbf{92.0} & 85.7 & \textbf{86.5} & \textbf{83.7} & \textbf{87.7} & 89.0 & 90.5 & \textbf{61.5} & \textbf{71.0} & \textbf{69.0} & \textbf{77.4} & \textbf{63.9} & \textbf{77.0} & \textbf{52.7} & \textbf{71.8} & \textbf{46.6} & \textbf{68.5} & 64.7 & 79.1 & \textbf{\meanvalue{70.7}} & \textbf{\meanvalue{80.1}} \\

\rowcolor{mygray}
\multicolumn{23}{c}{\textbf{\textit{(c) Training Stage}}} \\
\midrule
One-stage (Only RL)
& 89.5 & 91.7 & \textbf{85.8} & {86.5} & \textbf{84.9} & 87.4 &\textbf{90.5}& \textbf{91.7} & \textbf{64.2} & 70.6 & {69.0} & 75.0 & 56.7 & 71.7 & 36.0 &61.1  & 44.2 & 67.5 & 53.1 & 72.7   & \meanvalue{67.4} & \meanvalue{77.6} \\
\rowcolor{Light}
Two-stage (SFT + RL)
& \textbf{89.9} & \textbf{92.0} & 85.7 & \textbf{86.5} & 83.7 & \textbf{87.7} & 89.0& 90.5 & 61.5 & \textbf{71.0} & \textbf{69.0} & \textbf{77.4} & \textbf{63.9} & \textbf{77.0} & \textbf{52.7} & \textbf{71.8} & \textbf{46.6} & \textbf{68.5} & \textbf{64.7} & \textbf{79.1} & \textbf{\meanvalue{70.7}} & \textbf{\meanvalue{80.1}} \\
\bottomrule
\end{tabular}
        \end{adjustbox}
    \end{minipage}
\end{table*}

\subsection{Ablation Study}
As shown in \cref{tab:overall_ablation}, we conduct ablation experiments to analyze contributions of key designs in ConceptSeg-R1. See \cref{appendix:Reward},~\ref{appendix:Router},~\ref{appendix:Translation},~\ref{appendix:training_random_prompt} for more results.

\textit{\textbf{Architecture Evolution.}}
We first establish a strong baseline by introducing an MLLM with supervised fine-tuning (SFT), where the model generates geometric bounding boxes as intermediate prompts for SAM 3. This simple integration already achieves performance that exceeds competing methods in \cref{tab:quantitative_comparison}, ensuring subsequent design choices lead to convincing performance improvements.
Building upon this baseline, introducing GRPO yields a modest improvement in mean {\tt mIoU} (70.1 → 70.5), suggesting reinforcement learning improves reasoning stability. Replacing geometric prompts with the Concept Translation Module (CTM) leads to a substantial gain (70.5 → 75.8), highlighting the benefit of translating reasoning states into dense concept representations. The full Meta-GRPO configuration further improves performance to 80.1 mean {\tt mIoU}, demonstrating the proposed design choices contribute consistent and convincing performance improvements across diverse concepts.

\textit{\textbf{Meta-GRPO Component Analysis.}}
We further analyze the mechanisms of Meta-GRPO by removing key components. Replacing the Meta-Reward  with a standard bounding box {\tt IoU} reward leads to clear performance degradation on out-of-distribution datasets, indicating that without the meta-reward constraint, the model tends to memorize training cases rather than induce transferable task rules. 
Similarly, bypassing the meta-reasoning process results in a noticeable drop in accuracy, particularly on CR concepts, where rule verification is essential for resolving complex visual-textual ambiguities. Overall, removing either the meta-reward or the meta-reasoning component consistently reduces performance, demonstrating that both reward design and reasoning verification are necessary for stable rule induction and robust generalization.

\textit{\textbf{Training Strategy.}}
We further compare one-stage reinforcement learning with the proposed two-stage training scheme. The two-stage setting achieves higher overall performance and more stable results across CI, CD, and CR concepts, confirming the importance of initializing the model with supervised instruction alignment before reinforcement learning.

\section{Conclusion}
\label{sec:conclusion}

We present ConceptSeg-R1, a unified framework for generalized concept segmentation across context-independent (CI), context-dependent (CD), and context-reasoning (CR) concepts. By introducing rule-induced concept grounding and reference-conditioned reasoning, ConceptSeg-R1 enables segmentation to adapt to diverse contextual and reasoning requirements beyond static category recognition. Extensive experiments demonstrate strong performance across a wide range of natural, industrial, medical, and reasoning-intensive scenarios, while preserving native capability and efficiency of promptable segmentation backbones.
As an initial step toward segmenting any concept, ConceptSeg-R1 provides a practical baseline for advancing segmentation from object-level prediction toward concept-level understanding. We hope this work will encourage future research on unified concept segmentation, particularly in more complex multi-modal and real-world settings.

\clearpage
\appendix

\section*{Appendix}

\startcontents[sections]
\printcontents[sections]{l}{1}{\setcounter{tocdepth}{2}}

\clearpage

\section{Concept Definition}
\label{appendix:Concept_Definition}

We provide formal definitions and boundary conditions for the three levels of concept complexity used throughout the paper: context-independent (CI), context-dependent (CD), and context-reasoning (CR) concepts. These definitions aim to clarify the semantic scope of each concept category  
by characterizing the source of concept identity: intrinsic attributes for CI concepts, contextual relations for CD concepts, and higher-order reasoning structures for CR concepts.

\subsection{Context-Independent Concepts (CI)}

A concept is defined as context-independent if its identity can be determined primarily by intrinsic visual appearance and semantic attributes, without requiring explicit reasoning about surrounding context or relationships. Typical examples include common object classes, man-made artifacts, and fine-grained biological species. CI concepts correspond to appearance-driven recognition and represent the baseline level of perception in generalized segmentation.
Formally, a segmentation target belongs to the CI category if the mapping between an input image $x$ and its segmentation output $y$ can be determined from a single observation:
\begin{equation}
y = f_{\mathrm{CI}}(x)
\end{equation} 
In this setting, the target concept has a self-contained semantic identity.
The prediction does not depend on external reference images, relational constraints, or logical inference beyond appearance-based recognition.

\subsection{Context-Dependent Concepts (CD)}
A concept is defined as context-dependent if its identification requires understanding relationships between the target and its surrounding environment. These relationships may involve foreground-background contrast, transparency, occlusion, or domain-specific context such as industrial anomalies or medical lesions. In contrast to CI concepts, the identity of the target cannot be determined solely from intrinsic appearance, but instead emerges from its interaction with environmental context.
Representative examples of CD concepts include
saliency, where the target stands out from the background;
camouflage, where the target blends into its surroundings;
transparency or shadows, where the target appearance is altered by contextual visual interactions;
industrial anomalies, whose definition depends on manufacturing conditions and requirements;
and medical lesions, which must be interpreted relative to anatomical context.
Formally, the segmentation of a CD concept depends on the contextual structure in which the concept is semantically situated:
\begin{equation}
    y = f_{\mathrm{CD}}(x, \mathcal{C}_x)
\end{equation}
where $\mathcal{C}_x$ denotes contextual information within the image, such as background structure, spatial relationships, or domain-specific conditions.
The role of $\mathcal{C}_x$ is not merely auxiliary.
It is part of the semantic definition of the concept itself.
This formulation emphasizes that the segmentation decision relies on relational perception rather than isolated object recognition. 

\subsection{Context-Reasoning Concepts (CR)}
A concept is defined as context-reasoning if its semantic identity is specified by higher-order reasoning structures, such as cross-instance correspondence, scene differences, temporal relations, or other rules derived from reference information.
Unlike CD concepts, whose identities arise from local or domain-specific relations within an environment, CR concepts are defined by abstract relational rules that may span multiple observations, modalities, time steps, or logical constraints.
Typical reasoning scenarios include
consistency reasoning to establish correspondences across multiple images,
difference reasoning to identify changes between scenes,
logical reasoning to determine functional relationships,
and spatio-temporal reasoning to track objects across time or viewpoints.
These represent the higher level of cognitive complexity in the proposed taxonomy because their identities are not directly tied to fixed appearance or local context, but to inferred relational structures.
Formally, the semantic specification of a CR concept can be expressed as a two-step process of rule abstraction and application:
\begin{align}
    \gamma & = \Psi(m_1, m_2, \ldots, m_k) \\
    y      & = f_{\mathrm{CR}}(x, \gamma) \label{equ:general_concept_defination}
\end{align}
where $\{m_1, m_2, \ldots, m_k\}$ denotes a set of mixed-type reference inputs  (such as support images and guiding texts).
$\Psi(\cdot)$ denotes the reasoning operator that abstracts these references into a defining rule $\gamma$, and $f_{\mathrm{CR}}(\cdot)$ applies this rule on the target input $x$.
This formulation explicitly captures the rule-defined nature of CR concepts.

\subsection{Unified Concept Definition from the CR Perspective} 
From the perspective of CR concepts, the preceding CI, CD, and CR formulations can be unified by treating $\gamma$ as the variable component that determines the source of concept identity.
Specifically, the general rule-based formulation in~\cref{equ:general_concept_defination} can be extended to all three concept levels by defining $\gamma$ as follows:
\begin{equation}
    \gamma =
    \begin{cases}
        \emptyset                   & \text{for CI concepts} \\
        \mathcal{C}_x               & \text{for CD concepts} \\
        \Psi(m_1, m_2, \ldots, m_k) & \text{for CR concepts}
    \end{cases}
    \Rightarrow
    y = 
    \begin{cases}
        f_{\mathrm{CI}}(x, \emptyset) \\
        f_{\mathrm{CD}}(x, \mathcal{C}_x) \\
        f_{\mathrm{CR}}\bigl(x, \Psi(m_1, m_2, \ldots, m_k)\bigr)    
    \end{cases}
\end{equation}
Here, $\emptyset$ indicates that no additional defining rule is introduced.

\subsection{Prompt-Based Instantiation of Concept Rules}
The unified formulation above provides a direct motivation for the prompt-based design of our model. While $\gamma$ specifies the intrinsic concept source of semantic identity, our model instantiates it dynamically through a prompt-conditioned rule $\eta$ derived from visual references and textual guidance. As stated in~\cref{sec:meta_grpo}, given a target image $\mathcal{I}$ and a textual prompt $\mathcal{T}$, we construct a visual reference set $\mathcal{R}$.
The concept rule is then obtained by:
\begin{align}
    \eta & = \Psi(\mathcal{R}, \mathcal{T}) \\
    y    & = f(\mathcal{I}, \eta)
\end{align}
where $\Psi(\cdot)$ instantiates the prompt-conditioned rule from the reference inputs, and $f(\cdot, \eta)$ applies this rule to the target image. 
A critical question arises: \textit{How can a unified framework accommodate such divergent abstraction levels, ranging from isolated appearance filtering to complex logical deduction?}
This is achieved by formulating textual instructions and visual references as complementary sources for concept rule induction.

Instead of pre-defining separate relational structures for different concept categories, the same formulation $\eta=\Psi(\mathcal{R}, \mathcal{T})$ allows the induced rule $\eta$ to vary according to the semantic or reasoning demand specified by $\mathcal{T}$.
Based on the aforementioned concept definitions, this prompt-based formulation therefore manifests as three distinct rule induction demands under a unified framework.
Here, $\leadsto$ denotes the expected implicit abstraction behavior induced by the unified operator $\Psi(\cdot)$, rather than an explicitly pre-constructed input structure. 
The symbol $\leftrightarrow$ is used to indicate the intended semantic interaction or relational dependency between different sources, rather than a specific architectural operation.
\begin{itemize}[leftmargin=*,itemsep=0em,topsep=0em,parsep=0em]
    \item For CI concepts, the rule induction is strictly object-centric and focus on object-level appearance abstraction.
    The textual instruction $\mathcal{T}^{\mathrm{sem}}$ provides the semantic anchor, while the reference objects $\{o_i\}_{i=1}^{N}$ from visual references $\mathcal{R}=\{v_i\}_{i=1}^{N}$ provide appearance evidence of the target concept.
    Under this demand, the induced rule should emphasize the shared intrinsic appearance of the target objects:
    \begin{equation}
        \eta_{\mathrm{CI}} = \Psi_{\mathrm{CI}}(\mathcal{R}, \mathcal{T}^{\mathrm{sem}})
        \;\leadsto\;
        \mathcal{T}^{\mathrm{sem}} \leftrightarrow \{o_i\}_{i=1}^{N}
    \end{equation}

    \item For CD concepts, the rule induction shifts from isolated object appearance to intra-sample contextual abstraction.
    The semantic instruction $\mathcal{T}^{\mathrm{sem}}$ acts as a relational query that specifies the contextual pattern to be identified.
    Under this demand, the induced rule should capture how the target foreground $o_i$ is defined through its interaction with the environmental context $\mathcal{C}_{o_i}$, and distill the invariant contextual dependency across references:
    \begin{equation}
        \eta_{\mathrm{CD}} = \Psi_{\mathrm{CD}}(\mathcal{R}, \mathcal{T}^{\mathrm{sem}})
        \;\leadsto\;
        \mathcal{T}^{\mathrm{sem}} \leftrightarrow \{\,o_i \leftrightarrow \mathcal{C}_{o_i}\,\}_{i=1}^{N}
    \end{equation}
    
    \item For CR concepts, the rule induction further shifts toward complex relational rule deduction across the entire reference set.
    The reasoning instruction $\mathcal{T}^{\mathrm{rea}}$ specifies the task demand and conceptual premise of the target rule.
    Under this demand, the induced rule should be derived from the joint interpretation of the reasoning instruction and the available visual references $\{v_i\}_{i=1}^{N}$:
    \begin{equation}
        \eta_{\mathrm{CR}} = \Psi_{\mathrm{CR}}(\mathcal{R}, \mathcal{T}^{\mathrm{rea}})
        \;\leadsto\;
        \mathcal{T}^{\mathrm{rea}} \leftrightarrow \bigl( v_1 \leftrightarrow v_2 \leftrightarrow \dots \leftrightarrow v_N \bigr)
    \end{equation}
    When multiple visual references are available, this rule induction may involve inter-sample relational reasoning aligned with the reasoning instruction.
    When only a single reference image is available, the inter-sample relation naturally degenerates, and the rule is constructed primarily from the text-defined reasoning premise while being grounded by the available visual evidence.
\end{itemize}
This framework explains how the identical cross-modal architecture can adapt its rule induction behavior under different concept complexities.
It shifts fluidly from object-level appearance abstraction (CI), to intra-sample contextual abstraction (CD), and ultimately to complex relational rule deduction (CR), thereby supporting a unified modeling of highly heterogeneous concept categories.

\begin{figure*}[!t]
  \centering
  \includegraphics[width=1\linewidth]{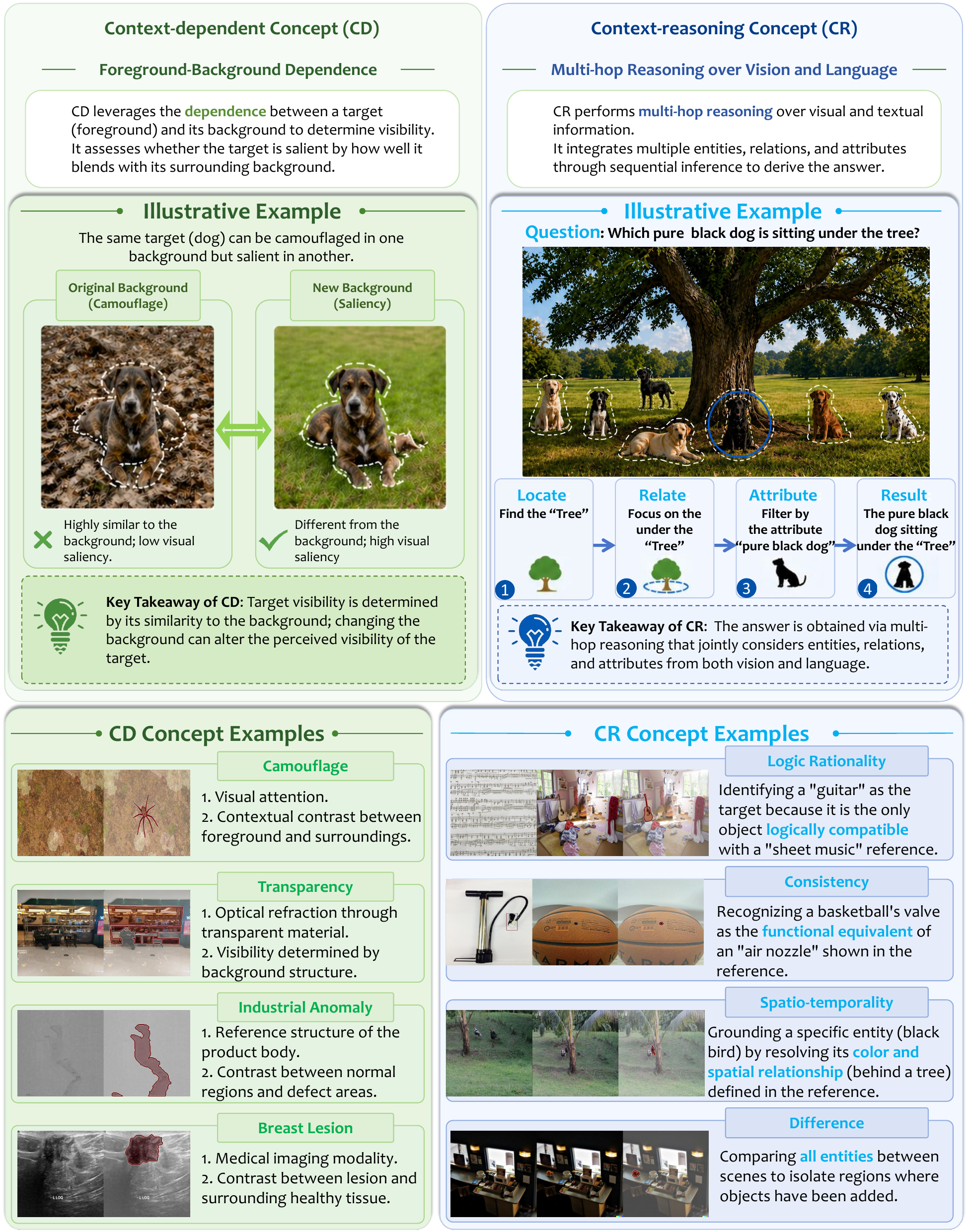}
  \caption{Visual comparison between CD and CR Concepts.}
  \label{fig:Concept_explaination}
\end{figure*}

\subsection{Visualizing Dependency and Reasoning Conditions}
CI concepts, such as people, vehicles, or ships, are relatively stable and environment-invariant, making their identities straightforward to determine from intrinsic appearance alone. 
In contrast,  CD or CR concepts require additional forms of dependency beyond object identity. 
As shown in Fig.~\ref{fig:Concept_explaination}, several representative CD and CR concepts are illustrated together with the conditions that distinguish them as different concept categories.
CD concepts primarily rely on \emph{foreground-background dependence}, where the target's visibility and concept identity are determined by its relationship to the surrounding context.
For instance, camouflage is defined by visual similarity between the target and its background, transparency depends on background refraction and scene composition, industrial anomalies are determined through comparison with the structural regularity of the product body, and medical lesions become identifiable only under imaging conditions relative to surrounding healthy tissue.
These examples demonstrate that CD targets often lack clear semantic meaning in isolation and can be reliably identified only through contextual contrast.
CR concepts require \emph{multi-hop reasoning} across entities, relations, and attributes.
Identification is achieved through a sequential inference process that integrates visual and textual cues.
For instance, locating a ``pure black dog sitting under the tree'' requires first grounding the tree, establishing the spatial relation ``under'', and then filtering by the attribute ``pure black''. 
Similarly, tasks such as logical compatibility, functional equivalence, or temporal consistency depend on reasoning over relationships rather than visual contrast alone.

\section{Hierarchical Concept Segmentation Benchmark Suite}
\label{appendix:Benchmark}

We curate a hierarchical concept segmentation benchmark suite to evaluate the out-of-distribution generalization of ConceptSeg-R1 across CI, CD, and CR concepts.
As summarized in \cref{tab:datasets}, this suite integrates a series of existing benchmarks and is organized into a training split for rule induction and an evaluation split structured according to the proposed CI/CD/CR taxonomy.
In addition, we introduce supplementary evaluation benchmarks in \cref{tab:add_benchmarks}.

\begin{table*}[t]
    \centering
    \caption{Concept segmentation dataset overview.
      ``*'' denotes re-organized or modified datasets (\cref{appendix:Benchmark}).}
    \label{tab:datasets}
    \begin{adjustbox}{width=\linewidth}
        
\begin{tabular}{
  l  
  r  
  r  
  r  
  r  
}

  & \textbf{Concept}
  & \textbf{Dataset}
  & \textbf{\#Images} 
  & \textbf{Description} \\
\toprule

\rowcolor{mygray}
\multicolumn{5}{c}{\textit{\textbf{Training Split}} \quad {\footnotesize (Total: 42,860 images)}} \\
\midrule

\multirow{2}{*}{CD}
  & Saliency & DUTS-TR~\cite{wang2017learning} & 10,548 & Large-scale salient object, diverse natural scenes \\
  & Camouflage & COD10K-TR~\cite{fan2020camouflaged} & 4,040   & Camouflaged animals/objects in natural habitats \\

\cmidrule{2-5}
\multirow{4}{*}{CR}
  & \multirow{3}{*}{Consistency} & FSS-1000R~\cite{li2020fss} & 10,000 & Few-shot segmentation, diverse object classes \\
  & & MGrounding*~\cite{li2025migician} & 9,102   & Reasoning to identify co-existing objects across images \\
  & & CoSOD3K~\cite{fan2021re} & 3,316  & Co-salient objects in relevant image groups \\
  & Difference & MGrounding*~\cite{li2025migician} & 5,854 & Reasoning to identify distinct objects between images \\

\midrule
\rowcolor{mygray}
\multicolumn{5}{c}{\textit{\textbf{Evaluation Benchmark}} \quad {\footnotesize (Total: 29,696 images)}} \\
\midrule

\multirow{4}{*}{CI}
 & Living Classes   & COCO20i*~\cite{shaban2017one} & 1,700 (17 Classes)  & Common living object classes for base segmentation \\
 & Artifact Classes   & COCO20i*~\cite{shaban2017one} & 6,244  (63 Classes) & Man-made artifact categories in complex scenes \\
  & Fine-grained Classes & iNaturalist*~\cite{carion2025sam} & 5,325 (2,537 Classes)  & Fine-grained species recognition localization \\
  & Ultra Rare Classes & iNaturalist*~\cite{carion2025sam} & 52  (26 Classes) & Long-tail distribution, extremely rare species \\
\cmidrule{2-5}
\multirow{7}{*}{CD} 
  & Optical Property (Transparency) & Trans10K~\cite{xie2020segmenting} & 4,428 &  Transparent object segmentation in real scenes \\
  &Optical Property (Shadow) & SBU~\cite{vicente2016large} & 638  & Shadow detection and reflection surface analysis \\
  & Camouflage & COD10K-TE~\cite{fan2020camouflaged} & 2,026  & Held-out camouflaged object evaluation split \\
  & Saliency & DUTS-TE~\cite{wang2017learning} & 5,017 &  Standard salient object detection test split \\
  & Industrial Anomaly & ESDIs-SOD*~\cite{cui2023autocorrelation} & 717 (12 classes) & Surface defects in diverse industrial materials \\
  & Polyp Lesion & Colon Polyp~\cite{fan2020pranet} & 798  & Clinical colonoscopy images for polyp detection \\
  & Breast Lesion& BUSI~\cite{al2020dataset} & 161  & Breast Ultrasound images (Normal, Benign, Malignant) \\
  &Skin Lesion & ISIC18~\cite{codella2019skin} & 808 & Skin lesion analysis and dermoscopy segmentation \\
\cmidrule{2-5}
\multirow{8}{*}{CR}
  & Common Consistency    & MIG*~\cite{li2025migician} & 477 & Identify shared concepts among diverse images \\
  & Correspondence Consistency & MIG*~\cite{li2025migician} & 84 & Semantic matching and pixel-level correspondence \\
  & Reference Consistency      & MIG*~\cite{li2025migician} & 93 & Locate targets based on specific reference prompts \\
  & Static Difference   & MIG*~\cite{li2025migician} & 187 & Discriminate varied objects from reference set \\
  & View Difference   & MIG*~\cite{li2025migician} & 85 & Identify changes across different viewpoints \\
  & Logical Rationality & MIG*~\cite{li2025migician} & 89 & High-level logical inference and attribute mapping \\
  & Cross-View  Spatio-temporality & MIG*~\cite{li2025migician} & 224 & Consistency across multi-angle scene captures \\
  & Cross-Frame Spatio-temporality  & MIG*~\cite{li2025migician} & 543 & Dynamic reasoning for temporal object persistence \\

\bottomrule
\end{tabular}
    \end{adjustbox}
\end{table*}

\begin{table*}[t]
    \centering
    \caption{Additional evaluation benchmarks. Generic segmentation benchmarks evaluate whether ConceptSeg-R1 preserves the native segmentation capability inherited from SAM 3, while reasoning-oriented benchmarks assess external reasoning-grounded segmentation ability.}
    \label{tab:add_benchmarks}
    \begin{adjustbox}{width=0.9\linewidth}
        \begin{tabular}{lr r r r}
   \textbf{Concept} 
    & 
    \textbf{Task} 
    & \textbf{Dataset} 
    & \textbf{\#Images} 
    & \textbf{Purpose} \\
    \toprule
    CI &
    Urban Scene Semantic Segmentation
    & Cityscapes~\cite{Cityscapes}
    & 500 (19 Classes) 
    & Scene-level robustness \\
    
    \midrule
      \multirow{2}{*}{CR} 
      &
   \multirow{2}{*}{Reasoning Segmentation}
    & ReasonSeg-Val~\cite{LISA}
    & 200 
    & Evaluation of basic reasoning logic \\
     &
    & ReasonSeg-Test~\cite{LISA}
    & 779
    &  Intricate reasoning and world knowledge \\
    \bottomrule
\end{tabular}
    \end{adjustbox}
\end{table*}

\textit{\textbf{Training Sets.}} 
The training split focuses on CD and CR concepts, since SAM 3 already provides strong native capability for many CI concepts. 
For CD concepts, we use salient object detection and camouflaged object detection as a reciprocal pair to strengthen foreground-background contrast perception. 
For CR concepts, we use consistency reasoning and difference reasoning datasets to encourage the model to distinguish shared patterns from discriminative variations across images. 
This design encourages ConceptSeg-R1 to learn transferable rules while preserving the inherent segmentation ability of the SAM 3 backbone.

\textit{\textbf{Dataset Reconstruction.}}
We detail the reconstruction and modification of existing datasets (``*'' in \cref{tab:datasets}) to construct our benchmark:
\begin{itemize}[leftmargin=*,itemsep=0em,topsep=0em,parsep=0em]
\item COCO20i~\cite{shaban2017one}: We manually partition data classes into \textit{Artifact and Living Classes}. For each class, we randomly sample up to 100 images. For the classes with fewer than 100 images, all available samples are included. 
\item iNaturalist~\cite{carion2025sam}: Classes are stratified into \textit{Fine-grained and Ultra Rare Classes} based on SAM 3's prediction confidence. The bottom 1\% of classes by confidence are designated as Ultra Rare. For each class, we sample 10\% of the images, ensuring a minimum of two images per class. 
\item ESDIs-SOD~\cite{cui2023autocorrelation}: We preserve most classes with precise annotations and discard the 3$^{rd}$ and 5$^{th}$ classes due to their diffuse and loosely distributed labeled areas, which do not meet our quality standards. 
\item MGrounding~\cite{li2025migician}: We repurpose the MGrounding dataset by elevating its bounding box annotations to pixel-level masks via SAM 3.
\item MIG~\cite{li2025migician}: Based on the multi-task grounding framework in MIG, we convert bounding boxes into high-quality masks via SAM 3 and manually filter out low-quality samples. A rigorous manual audit was conducted to prune instances with ambiguous boundaries or low-quality masks, ensuring a clean and dense supervision signal for the concept query learning. 
To construct our reasoning-heavy benchmark, we systematically adapted the original MIG grounding tasks into eight segmentation sub-tasks:
    \begin{enumerate}[leftmargin=*,itemsep=0em,topsep=0em,parsep=0em]
        \item \textit{Common Consistency} from Common Object Grounding; 
        \item \textit{Correspondence Consistency} from Correspondence Grounding; 
        \item \textit{Reference Consistency} from Referring Grounding; 
        \item \textit{Static Difference} from Static Difference Grounding; 
        \item \textit{View Difference} from View Difference Grounding; 
        \item \textit{Logical Rationality} from Reasoning Grounding; 
        \item \textit{Cross-View Spatio-temporality} from Multi-View Grounding; 
        \item \textit{Cross-Frame Spatio-temporality} from Object Tracking.
    \end{enumerate}
\end{itemize} 

\textit{\textbf{Evaluation Benchmark.}}
The main evaluation benchmark in \cref{tab:datasets}  is organized into three levels of cognitive complexity to decouple intrinsic recognition from context-aware and reasoning-intensive segmentation. 
CI concept segmentation tasks evaluate appearance-driven recognition, including common classes, artifact classes, fine-grained classes, and ultra-rare long-tail concepts. 
CD concept segmentation tasks evaluate optical properties, camouflage, saliency, industrial anomalies, and medical lesions, where targets are defined by their relations to the environment. 
CR concept segmentation tasks evaluate multi-step visual-textual reasoning, including consistency, difference, logical rationality, and spatio-temporality, requiring cross-image dependencies or higher-order relational understanding. 
This hierarchy provides a unified protocol for assessing generalized concept segmentation across perception, contextual understanding, and reasoning.

\textit{\textbf{Supplementary Evaluation.}}
As shown in \cref{tab:add_benchmarks}, we further introduce several external benchmarks to provide a broader evaluation of ConceptSeg-R1.
These benchmarks can still be interpreted within our concept framework, but we report them in their original benchmark form rather than further subdividing them by concept category, so as to examine performance preservation and facilitate comparison with existing methods under existing used evaluation protocols.
Specifically, in~\cref{tab:cityscapes}, we evaluate the model on Cityscapes~\cite{Cityscapes}, a generic segmentation benchmark that conforms to our CI concept definition, as its semantic categories are primarily determined by stable visual semantics. 
This evaluation verifies that reasoning-oriented training does not compromise the general-purpose, CI-style segmentation ability inherited from SAM 3.
We also include ReasonSeg-Val/Test~\cite{LISA} as an external reasoning-grounded segmentation benchmark in~\cref{tab:zeroshot_reasonseg}.
ReasonSeg mainly evaluates single-image reasoning-grounded segmentation with textual instructions, and can be viewed as a CR-style evaluation closely related to logical rationality.

\begin{table}[t!]
    \centering
    \caption{Hyper-parameters configuration.}
    \label{tab:sft_rl_config}
    \subfloat[Stage 1: Supervised Fine-Tuning.]{\label{tab:sft_config}
        \centering
        \begin{adjustbox}{width=0.5\linewidth}
            
\begin{tabular}{lr}

\textbf{Configuration} & \textbf{Value} \\ 
\toprule
Training Epochs & 20 \\
Batch Size & 128 \\
SAM 3 Image Size & $1008 \times 1008$ \\
Learning Rate & $3 \times 10^{-5}$ \\
Optimizer & AdamW \\
Learning Rate Scheduler & Linear \\
Max Prompt Length & 2048 \\
Max Completion Length & 768 \\
Concept Query Length ($L_2$) & 8 \\
GRPO Beta ($\beta$) & 0.04 \\
Trainable Parameters & CTM \\
Random Seed & 42 \\ 
\bottomrule
\end{tabular}

        \end{adjustbox}
    }
    \subfloat[Stage 2: Cognitive Fine-Tuning.]{\label{tab:rl_config}
        \centering
        \begin{adjustbox}{width=0.48\linewidth}
            
\begin{tabular}{lr}
\
\textbf{Configuration} & \textbf{Value} \\ 
\toprule
Training Epochs & 2 \\
Batch Size & 64 \\
SAM 3 Image Size & $1008 \times 1008$ \\
Learning Rate & $1 \times 10^{-6}$ \\
Optimizer & AdamW \\
Learning Rate Scheduler & Linear \\
Max Prompt Length & 2048 \\
Max Completion Length & 768 \\
Concept Query Length ($L_2$) & 8 \\
GRPO Group Size ($G$) & 8 \\
GRPO Beta ($\beta$) & 0.04 \\
Trainable Parameters & MLLM, CTM \\
Random Seed & 42 \\ 
\bottomrule
\end{tabular}
    
        \end{adjustbox}
    }

\end{table}

\section{Implementation Details}
\label{appendix:Implementation} 
Our training process consists of two primary stages: supervised fine-tuning for alignment and cognitive reinforcement learning for reasoning optimization.
All experiments are conducted using the AdamW~\cite{adamw} optimizer with a linear learning rate scheduler, where reference and reasoning images are uniformly resized to $600 \times 600$ pixels.
To ensure reproducibility, the random seed is fixed at 42.

\subsection{Stage 1: Supervised Fine-Tuning}

In the first stage, we focus on aligning the multi-modal reasoning capabilities with the concept translation module (CTM).
We employ an image resolution of $1008 \times 1008$ for the SAM 3 backbone and $600 \times 600$ for reference images.
During this stage, only the parameters of the CTM are updated.
Detailed hyper-parameters are provided in \cref{tab:sft_config}.

\subsection{Stage 2: Cognitive Reinforcement Learning}

In the second stage, we utilize Meta-GRPO to enhance the model's deductive reasoning and rule-following abilities. The learning rate is reduced to $1 \times 10^{-6}$ to maintain training stability. We jointly optimize the MLLM and the CTM. For each prompt, we sample a group of $G=8$ outputs to calculate the relative advantage within GRPO. The detailed configurations are summarized in \cref{tab:rl_config}. 

\section{Prompt Templates}
\label{appendix:Templates}

In this section, we detail the structured prompt templates employed in ConceptSeg-R1. To bridge the gap between low-level visual features and high-level conceptual reasoning, we adopt a hierarchical instruction format that organizes input signals into \textit{Support}, \textit{Proxy}, and \textit{Query} components.

\subsection{Standard Instruction Template}
Our template is designed to elicit explicit chain-of-thought (CoT) rationales before generating spatial coordinates. The system prompt enforces a strict rule-induction protocol, requiring the model to verify induced rules on a specific sub-region (the ``Check'' step) before finalizing the target localization.
\begin{mdframed}[
    linecolor=black!75,
    linewidth=1pt,
    backgroundcolor=gray!5,
    innertopmargin=10pt,
    innerbottommargin=10pt,
    innerleftmargin=10pt,
    innerrightmargin=10pt,
    roundcorner=2pt,
    frametitle={System Prompt \& Template},
    frametitlebackgroundcolor=gray!20
]
\small
\textbf{Instruction:}
Your task is to locate the object matching \texttt{{problem}} in the target image.

\textbf{Reference Image:}
Bounding boxes provided at \texttt{{reference\_boxes}}.
\\
\textbf{Target Image:}
The image for grounding.

\textbf{Protocol:} 
Think through the reasoning process in your mind, induce the visual rule and check it by locating the missed bounding box in the  reference image, apply this rule to locate the corresponding object in the Target Image.

\textbf{Output strictly in the following format:} 
<think>[Your step-by-step analysis and reasoning]</think>  <rule>Visual rule of the reference targets</rule>  <check>[x1, y1, x2, y2]</check> <bbox>[x3, y3, x4, y4]</bbox> <answer>concise noun phrase for target object</answer>
\end{mdframed}

\subsection{Token Representations and Semantic Placeholders}
To ensure the multi-modal large language model (MLLM) correctly interprets the input data, we define the following special tokens and placeholders:
\begin{itemize}[leftmargin=*,itemsep=0em,topsep=0em,parsep=0em]
    \item \texttt{[problem]}: A natural language query or conceptual description (e.g., ``the most fragile object'').
    \item \texttt{[reference\_boxes]}: Normalized coordinates $[x_{min}, y_{min}, x_{max}, y_{max}]$ representing the \textit{Support}   set targets.
    \item \texttt{<think>}: Dedicated tags for the Reasoning-Chain (CoT), incentivized via GRPO to perform rule induction.
    \item \texttt{<rule>}: A concise linguistic abstraction representing the relational mapping between the reference and target images. It encapsulates the underlying logic of the task (e.g., identity, difference, or functional scaling).
    \item \texttt{<check>}: A diagnostic bounding box used to verify the induced rule on the reference image's ``proxy query''.
    \item \texttt{<bbox>}: The final predicted coordinates for the Target Image.
    \item \texttt{<answer>}: A brief, 1-3 word noun phrase identifying the segmented concept to ensure linguistic grounding,  which are subsequently passed to the Segment Anything Model (SAM 3).
\end{itemize}

\subsection{Compositional Sensitivity and Mosaic Processing}
The template is highly adaptive to various input configurations and data augmentations:
\begin{itemize}[leftmargin=*,itemsep=0em,topsep=0em,parsep=0em]
    \item \textbf{Support/Proxy Scaling ($K^2$-shot):} The template dynamically accommodates $K^2$ support samples. As $K$ increases, the \texttt{[reference\_boxes]} list expands, facilitating complex cross-instance comparisons.
    \item \textbf{Reference Image Mosaic Strategy:} To handle multi-image context within a fixed window, reference images are organized into a $K \times K$ mosaic. Each image is resized to a specific sub-region (e.g., 1/2 size for a $2 \times 2$ grid). Original masks and bounding box coordinates undergo a corresponding scale transformation to fit the new mosaic layout. 
    \item \textbf{Global Coordinate  Normalization:} All spatial coordinates within the template are globally normalized relative to the entire mosaic canvas. This ensures a consistent spatial reference frame for rule induction and proxy verification across diverse aspect ratios.
    \item \textbf{Proxy Query Presence:} The template facilitates toggling the ``Proxy Query'' (i.e., the intentionally omitted bounding box in the reference image). When activated, it functions as an intermediate reasoning anchor for rule verification. In single-reference scenarios, the model directly induces the task rule and proceeds to localize the target object.
\end{itemize}

\begin{table*}[t]
    \centering
    \caption{Per-dataset results and mean performance.}
    \label{tab:quantitative_full}
    \begin{adjustbox}{width=\linewidth}
        \begin{tabular}{l*{23}{r}}
     & \multicolumn{5}{c}{\textit{CI Concepts}}
     & \multicolumn{9}{c}{\textit{CD Concepts}}
     & \multicolumn{9}{c}{\textit{CR Concepts}}                                                                                                                      \\
    \cmidrule(lr){2-6}
    \cmidrule(lr){7-15}
    \cmidrule(lr){16-24}
     & \makecell{\scriptsize Living\mcnl \scriptsize Classes}
     & \makecell{\scriptsize Artifact\mcnl \scriptsize Classes}
     & \makecell{\scriptsize Fine-grained\mcnl \scriptsize Classes}
     & \makecell{\scriptsize Ultra Rare\mcnl \scriptsize Classes}
     &
     & \multicolumn{2}{c}{\scriptsize Optical Property}
     & \multicolumn{1}{c}{\scriptsize Camouflage}
     & \multicolumn{1}{c}{\scriptsize Saliency}
     & \makecell{\scriptsize Industrial\mcnl \scriptsize Anomaly}
     & \makecell{\scriptsize Polyp\mcnl \scriptsize Lesion}
     & \makecell{\scriptsize Breast\mcnl \scriptsize Lesion}
     & \makecell{\scriptsize Skin\mcnl \scriptsize Lesion}
     &
     & \multicolumn{3}{c}{\scriptsize Consistency}
     & \multicolumn{2}{c}{\scriptsize Difference}
     & \multicolumn{1}{c}{\scriptsize Rationality}
     & \multicolumn{2}{c}{\scriptsize Spatio-temporality}
     &                                                                                                                                                      \\
    \cmidrule(lr){2-2}
    \cmidrule(lr){3-3}
    \cmidrule(lr){4-4}
    \cmidrule(lr){5-5}
    \cmidrule(lr){7-8}
    \cmidrule(lr){9-9}
    \cmidrule(lr){10-10}
    \cmidrule(lr){11-11}
    \cmidrule(lr){12-14}
    \cmidrule(lr){16-18}
    \cmidrule(lr){19-20}
    \cmidrule(lr){21-21}
    \cmidrule(lr){22-23}

    \textbf{Metric}
     & {\scriptsize COCO20i}
     & {\scriptsize COCO20i}
     & {\scriptsize iNaturalist}
     & {\scriptsize iNaturalist}
     & \textbf{Mean}
     & {\scriptsize SBU}
     & {\scriptsize Trans10K}
     & {\scriptsize COD10K-TE}
     & {\scriptsize DUTS-TE}
     & {\scriptsize ESDIs-SOD}
     & {\scriptsize Colon Polyp}
     & {\scriptsize BUSI}
     & {\scriptsize ISIC18}
     & \textbf{Mean}
     & {\scriptsize Common}
     & {\scriptsize Correspondence}
     & {\scriptsize Reference}
     & {\scriptsize Static}
     & {\scriptsize View}
     & {\scriptsize Logical}
     & {\scriptsize View}
     & {\scriptsize Frame}
     & \textbf{Mean}                                                                                                                                        \\
    \toprule

    \multicolumn{24}{c}{SAM 3~\cite{carion2025sam}}                                                                                                         \\
    \midrule
    {\tt MAE $\downarrow$}
     & 2.20                                                         & 1.98  & 0.50  & 0.57  & \meanvalue{1.31}
     & 6.10                                                         & 3.75  & 23.55 & 16.43 & 10.55             & 24.88 & 12.01 & 50.53 & \meanvalue{18.48}
     & --                                                           & --    & --    & --    & --                & --    & --    & --    & --                \\

    {\tt BER $\downarrow$}
     & 6.37                                                         & 7.93  & 1.70  & 1.87  & \meanvalue{4.47}
     & 13.03                                                        & 4.60  & 23.32 & 32.19 & 22.79             & 36.26 & 21.91 & 37.99 & \meanvalue{24.01}
     & --                                                           & --    & --    & --    & --                & --    & --    & --    & --                \\

    {\tt $F_{\beta}^{\omega} \uparrow$}
     & 86.28                                                        & 80.11 & 96.03 & 95.72 & \meanvalue{89.54}
     & 79.67                                                        & 91.46 & 51.28 & 38.06 & 51.48             & 17.46 & 49.57 & 41.86 & \meanvalue{52.60}
     & --                                                           & --    & --    & --    & --                & --    & --    & --    & --                \\

    {\tt $S_m \uparrow$}
     & 87.73                                                        & 82.14 & 93.50 & 95.91 & \meanvalue{89.82}
     & 79.30                                                        & 91.23 & 62.59 & 58.46 & 66.22             & 45.50 & 67.35 & 37.00 & \meanvalue{63.46}
     & --                                                           & --    & --    & --    & --                & --    & --    & --    & --                \\

    {\tt mIoU $\uparrow$}
     & 88.66                                                        & 85.40 & 96.74 & 96.47 & \meanvalue{91.82}
     & 81.49                                                        & 91.34 & 61.43 & 58.97 & 66.17             & 44.91 & 65.53 & 36.31 & \meanvalue{63.27}
     & --                                                           & --    & --    & --    & --                & --    & --    & --    & --                \\

    {\tt mDice $\uparrow$}
     & 87.23                                                        & 81.58 & 96.07 & 95.86 & \meanvalue{90.19}
     & 80.70                                                        & 92.87 & 54.44 & 39.70 & 53.11             & 21.61 & 53.79 & 48.34 & \meanvalue{55.57}
     & --                                                           & --    & --    & --    & --                & --    & --    & --    & --                \\

    \multicolumn{24}{c}{LENS-3B~\cite{zhu2026lens}}                                                                                                         \\
    \midrule
    {\tt MAE $\downarrow$}
     & 6.49                                                         & 4.16  & 1.23  & 0.96  & \meanvalue{3.21}
     & 15.25                                                        & 14.80 & 6.17  & 6.08  & 19.47             & 10.58 & 9.23  & 13.35 & \meanvalue{11.87}
     & 20.07                                                        & 10.36 & 8.33  & 14.01 & 5.89              & 6.90  & 4.92  & 3.12  & \meanvalue{9.20}  \\

    {\tt BER $\downarrow$}
     & 16.86                                                        & 17.00 & 2.49  & 2.32  & \meanvalue{9.67}
     & 32.33                                                        & 18.25 & 18.74 & 12.98 & 32.45             & 20.45 & 23.37 & 13.60 & \meanvalue{21.52}
     & 40.14                                                        & 29.65 & 33.31 & 35.60 & 32.53             & 40.86 & 37.01 & 14.88 & \meanvalue{33.00} \\

    {\tt $F_{\beta}^{\omega} \uparrow$}
     & 64.18                                                        & 55.70 & 89.98 & 87.74 & \meanvalue{74.40}
     & 43.20                                                        & 68.88 & 58.08 & 76.90 & 51.34             & 49.91 & 44.08 & 73.91 & \meanvalue{58.29}
     & 28.41                                                        & 20.87 & 30.77 & 22.28 & 28.98             & 19.52 & 25.73 & 61.61 & \meanvalue{29.77} \\

    {\tt $S_m \uparrow$}
     & 73.94                                                        & 71.81 & 92.23 & 90.83 & \meanvalue{82.20}
     & 58.15                                                        & 72.46 & 73.15 & 81.72 & 61.38             & 68.78 & 65.12 & 76.63 & \meanvalue{69.67}
     & 48.64                                                        & 53.93 & 58.94 & 52.17 & 59.93             & 53.54 & 58.03 & 77.82 & \meanvalue{57.88} \\

    {\tt mIoU $\uparrow$}
     & 74.29                                                        & 72.51 & 92.39 & 91.71 & \meanvalue{82.73}
     & 58.85                                                        & 71.91 & 73.08 & 81.05 & 60.17             & 67.72 & 65.00 & 74.68 & \meanvalue{69.06}
     & 51.08                                                        & 54.43 & 60.04 & 53.10 & 60.43             & 55.08 & 58.76 & 77.63 & \meanvalue{58.82} \\

    {\tt mDice $\uparrow$}
     & 65.58                                                        & 57.98 & 91.43 & 89.17 & \meanvalue{76.04}
     & 43.90                                                        & 70.24 & 60.42 & 77.15 & 50.38             & 53.42 & 48.55 & 75.24 & \meanvalue{59.91}
     & 28.18                                                        & 24.01 & 32.52 & 24.88 & 30.60             & 20.52 & 26.85 & 64.12 & \meanvalue{31.46} \\

    \multicolumn{24}{c}{Seg-Zero-7B~\cite{liu2025seg}}                                                                                                      \\
    \midrule
    {\tt MAE $\downarrow$}
     & 2.91                                                         & 2.57  & 0.62  & 0.49  & \meanvalue{1.65}
     & 13.50                                                        & 9.31  & 5.89  & 10.58 & 26.92             & 15.32 & 8.00  & 12.00 & \meanvalue{12.69}
     & 23.53                                                        & 7.86  & 8.17  & 7.63  & 3.67              & 9.17  & 6.13  & 7.99  & \meanvalue{9.27}  \\

    {\tt BER $\downarrow$}
     & 7.99                                                         & 10.57 & 1.97  & 1.39  & \meanvalue{5.48}
     & 26.32                                                        & 10.25 & 11.35 & 9.49  & 28.39             & 20.82 & 19.94 & 19.12 & \meanvalue{18.21}
     & 43.47                                                        & 44.71 & 42.99 & 46.42 & 48.83             & 46.11 & 47.37 & 48.23 & \meanvalue{46.02} \\

    {\tt $F_{\beta}^{\omega} \uparrow$}
     & 82.80                                                        & 73.28 & 95.10 & 96.03 & \meanvalue{86.80}
     & 54.33                                                        & 79.81 & 74.95 & 72.76 & 51.43             & 51.18 & 56.24 & 73.31 & \meanvalue{64.25}
     & 22.25                                                        & 7.20  & 18.89 & 9.32  & 1.72              & 9.80  & 7.50  & 6.80  & \meanvalue{10.44} \\

    {\tt $S_m \uparrow$}
     & 85.62                                                        & 79.19 & 95.31 & 93.58 & \meanvalue{88.43}
     & 63.33                                                        & 81.54 & 82.06 & 80.26 & 58.20             & 67.27 & 71.64 & 75.20 & \meanvalue{72.44}
     & 43.96                                                        & 47.47 & 52.34 & 47.79 & 47.16             & 47.88 & 48.03 & 46.95 & \meanvalue{47.70} \\

    {\tt mIoU $\uparrow$}
     & 86.29                                                        & 81.63 & 95.91 & 96.47 & \meanvalue{90.08}
     & 65.00                                                        & 81.21 & 81.11 & 78.56 & 57.12             & 65.72 & 70.70 & 73.07 & \meanvalue{71.56}
     & 45.74                                                        & 49.19 & 53.89 & 49.54 & 48.78             & 49.29 & 49.71 & 48.79 & \meanvalue{49.37} \\

    {\tt mDice $\uparrow$}
     & 83.81                                                        & 75.18 & 95.46 & 96.31 & \meanvalue{87.69}
     & 56.05                                                        & 82.95 & 76.82 & 76.34 & 52.36             & 54.29 & 59.77 & 72.80 & \meanvalue{66.42}
     & 24.45                                                        & 8.09  & 19.50 & 9.97  & 2.17              & 10.64 & 7.89  & 7.26  & \meanvalue{11.25} \\

    \multicolumn{24}{c}{SAM3-Agent-3B~\cite{carion2025sam}}                                                                                                 \\
    \midrule
    {\tt MAE $\downarrow$}
     & 7.33                                                         & 5.98  & 4.83  & 2.78  & \meanvalue{5.23}
     & 16.22                                                        & 16.75 & 11.52 & 9.61  & 17.19             & 14.84 & 23.57 & 51.88 & \meanvalue{20.20}
     & 16.47                                                        & 4.69  & 3.32  & 4.82  & 3.37              & 6.96  & 2.01  & 5.41  & \meanvalue{5.88}  \\

    {\tt BER $\downarrow$}
     & 26.56                                                        & 33.89 & 17.40 & 13.21 & \meanvalue{22.77}
     & 33.48                                                        & 24.57 & 29.37 & 23.35 & 45.22             & 46.66 & 41.40 & 58.46 & \meanvalue{37.81}
     & 34.68                                                        & 18.99 & 30.53 & 29.55 & 32.22             & 31.48 & 21.13 & 44.23 & \meanvalue{30.35} \\

    {\tt $F_{\beta}^{\omega} \uparrow$}
     & 45.11                                                        & 34.53 & 66.11 & 74.84 & \meanvalue{55.15}
     & 40.80                                                        & 54.17 & 45.36 & 59.47 & 20.59             & 10.06 & 19.92 & 21.23 & \meanvalue{33.95}
     & 25.78                                                        & 1.18  & 34.00 & 1.44  & 4.70              & 20.28 & 9.76  & 10.54 & \meanvalue{13.46} \\

    {\tt $S_m \uparrow$}
     & 62.60                                                        & 61.68 & 77.84 & 84.30 & \meanvalue{71.61}
     & 57.62                                                        & 65.94 & 64.91 & 72.08 & 49.20             & 47.24 & 46.71 & 25.40 & \meanvalue{53.64}
     & 50.13                                                        & 17.35 & 60.90 & 28.15 & 33.91             & 46.40 & 28.84 & 50.82 & \meanvalue{39.56} \\

    {\tt mIoU $\uparrow$}
     & 62.51                                                        & 62.05 & 79.91 & 84.92 & \meanvalue{72.35}
     & 58.22                                                        & 65.75 & 64.76 & 72.07 & 49.14             & 47.03 & 46.28 & 30.95 & \meanvalue{54.30}
     & 50.66                                                        & 17.75 & 61.04 & 28.03 & 33.81             & 46.90 & 29.03 & 51.92 & \meanvalue{39.89} \\

    {\tt mDice $\uparrow$}
     & 44.82                                                        & 34.38 & 65.89 & 74.67 & \meanvalue{54.94}
     & 40.98                                                        & 55.03 & 45.87 & 59.08 & 19.12             & 10.39 & 21.04 & 22.78 & \meanvalue{34.29}
     & 25.95                                                        & 1.48  & 33.94 & 1.38  & 4.53              & 20.48 & 9.75  & 11.20 & \meanvalue{13.59} \\

    \multicolumn{24}{c}{SAM3-Agent-7B~\cite{carion2025sam}}                                                                                                 \\
    \midrule
    {\tt MAE $\downarrow$}
     & 4.44                                                         & 3.35  & 0.98  & 0.97  & \meanvalue{2.44}
     & 10.09                                                        & 10.64 & 10.20 & 7.75  & 22.64             & 20.94 & 25.67 & 57.19 & \meanvalue{20.64}
     & 9.43                                                         & 2.65  & 3.15  & 5.63  & 2.82              & 3.75  & 0.71  & 7.22  & \meanvalue{4.42}  \\

    {\tt BER $\downarrow$}
     & 18.69                                                        & 20.96 & 7.26  & 4.73  & \meanvalue{12.91}
     & 23.66                                                        & 13.98 & 23.32 & 16.03 & 40.25             & 45.07 & 35.83 & 61.10 & \meanvalue{32.53}
     & 19.24                                                        & 20.78 & 31.45 & 28.37 & 31.34             & 28.62 & 12.19 & 38.22 & \meanvalue{26.28} \\

    {\tt $F_{\beta}^{\omega} \uparrow$}
     & 68.32                                                        & 60.60 & 86.31 & 91.79 & \meanvalue{76.76}
     & 61.33                                                        & 75.21 & 58.86 & 74.43 & 31.45             & 17.30 & 32.48 & 25.50 & \meanvalue{47.07}
     & 47.39                                                        & 1.38  & 31.37 & 25.05 & 28.29             & 29.25 & 13.76 & 13.16 & \meanvalue{23.71} \\

    {\tt $S_m \uparrow$}
     & 77.23                                                        & 73.45 & 88.72 & 93.36 & \meanvalue{83.19}
     & 68.84                                                        & 78.69 & 71.79 & 80.10 & 48.85             & 47.13 & 51.15 & 22.64 & \meanvalue{58.59}
     & 59.24                                                        & 19.97 & 59.78 & 48.52 & 56.63             & 52.78 & 23.35 & 49.51 & \meanvalue{46.22} \\

    {\tt mIoU $\uparrow$}
     & 77.46                                                        & 74.98 & 91.61 & 94.11 & \meanvalue{84.54}
     & 70.25                                                        & 78.44 & 71.65 & 80.20 & 49.03             & 46.91 & 50.70 & 29.93 & \meanvalue{59.64}
     & 60.00                                                        & 20.62 & 59.89 & 48.87 & 57.39             & 53.26 & 23.90 & 50.29 & \meanvalue{46.78} \\

    {\tt mDice $\uparrow$}
     & 67.88                                                        & 60.56 & 86.20 & 91.58 & \meanvalue{76.56}
     & 60.99                                                        & 76.44 & 59.25 & 74.04 & 30.36             & 17.97 & 33.42 & 26.88 & \meanvalue{47.42}
     & 47.65                                                        & 1.65  & 31.44 & 25.35 & 28.37             & 29.03 & 13.54 & 15.01 & \meanvalue{24.01} \\

    \rowcolor{Light}
    \multicolumn{24}{c}{ConceptSeg-R1-3B}                                                                                                                   \\
    \midrule
    {\tt MAE $\downarrow$}
     & 2.27                                                         & 2.17  & 0.40  & 0.49  & \meanvalue{1.33}
     & 6.10                                                         & 3.73  & 2.52  & 4.18  & 10.97             & 6.37  & 6.93  & 8.36  & \meanvalue{6.15}
     & 5.87                                                         & 11.41 & 1.99  & 9.95  & 2.36              & 6.55  & 4.31  & 1.35  & \meanvalue{5.47}  \\

    {\tt BER $\downarrow$}
     & 6.38                                                         & 8.13  & 1.03  & 1.63  & \meanvalue{4.29}
     & 13.03                                                        & 4.60  & 6.98  & 4.66  & 18.28             & 16.60 & 14.82 & 10.63 & \meanvalue{11.20}
     & 9.92                                                         & 18.33 & 10.14 & 21.24 & 19.66             & 28.06 & 26.93 & 5.84  & \meanvalue{17.52} \\

    {\tt $F_{\beta}^{\omega} \uparrow$}
     & 86.17                                                        & 79.93 & 96.92 & 96.45 & \meanvalue{89.87}
     & 79.82                                                        & 91.57 & 83.65 & 88.96 & 61.52             & 58.96 & 66.75 & 81.14 & \meanvalue{76.55}
     & 80.11                                                        & 34.59 & 77.07 & 49.68 & 55.78             & 46.57 & 49.90 & 79.51 & \meanvalue{59.15} \\

    {\tt $S_m \uparrow$}
     & 87.52                                                        & 81.80 & 93.65 & 96.07 & \meanvalue{89.76}
     & 79.22                                                        & 91.12 & 88.19 & 90.80 & 71.59             & 75.09 & 77.54 & 82.96 & \meanvalue{82.06}
     & 84.80                                                        & 60.03 & 85.53 & 67.34 & 73.14             & 67.49 & 69.62 & 87.49 & \meanvalue{74.43} \\

    {\tt mIoU $\uparrow$}
     & 88.55                                                        & 85.20 & 97.38 & 96.87 & \meanvalue{92.00}
     & 81.56                                                        & 91.37 & 87.65 & 90.47 & 70.97             & 74.31 & 76.33 & 81.50 & \meanvalue{81.77}
     & 85.77                                                        & 59.69 & 85.57 & 68.21 & 75.41             & 68.49 & 70.24 & 88.05 & \meanvalue{75.18} \\

    {\tt mDice $\uparrow$}
     & 87.09                                                        & 81.36 & 97.16 & 96.61 & \meanvalue{90.56}
     & 80.84                                                        & 92.94 & 84.87 & 89.74 & 62.57             & 61.40 & 69.92 & 81.55 & \meanvalue{77.98}
     & 81.39                                                        & 38.67 & 78.24 & 51.85 & 56.80             & 47.03 & 49.95 & 81.83 & \meanvalue{60.72} \\

    \rowcolor{Light}
    \multicolumn{24}{c}{ConceptSeg-R1-7B}                                                                                                                   \\
    \midrule
    {\tt MAE $\downarrow$}
     & 2.17                                                         & 2.19  & 0.39  & 0.43  & \meanvalue{1.30}
     & 6.06                                                         & 3.74  & 2.18  & 1.92  & 9.35              & 6.11  & 5.87  & 7.58  & \meanvalue{5.35}
     & 4.80                                                         & 5.23  & 1.65  & 3.97  & 2.18              & 3.22  & 3.01  & 1.01  & \meanvalue{3.13}  \\

    {\tt BER $\downarrow$}
     & 6.46                                                         & 8.68  & 0.98  & 1.09  & \meanvalue{4.30}
     & 13.00                                                        & 4.61  & 6.93  & 2.78  & 17.59             & 15.10 & 15.42 & 10.40 & \meanvalue{10.73}
     & 8.23                                                         & 19.47 & 11.10 & 16.26 & 24.33             & 19.38 & 21.79 & 5.48  & \meanvalue{15.82} \\

    {\tt $F_{\beta}^{\omega} \uparrow$}
     & 86.86                                                        & 79.39 & 96.56 & 96.76 & \meanvalue{89.89}
     & 79.84                                                        & 91.57 & 84.75 & 92.65 & 66.67             & 65.01 & 69.56 & 82.22 & \meanvalue{79.03}
     & 83.42                                                        & 49.81 & 77.13 & 64.98 & 48.99             & 60.22 & 58.12 & 81.39 & \meanvalue{65.51} \\

    {\tt $S_m \uparrow$}
     & 87.63                                                        & 81.73 & 96.05 & 93.62 & \meanvalue{89.76}
     & 79.22                                                        & 91.11 & 88.75 & 93.65 & 74.47             & 78.50 & 79.39 & 83.93 & \meanvalue{83.63}
     & 86.99                                                        & 68.03 & 85.25 & 76.41 & 70.18             & 76.27 & 73.96 & 88.75 & \meanvalue{78.23} \\

    {\tt mIoU $\uparrow$}
     & 88.77                                                        & 85.32 & 97.19 & 97.24 & \meanvalue{92.13}
     & 81.57                                                        & 91.37 & 88.28 & 93.53 & 74.10             & 77.20 & 78.17 & 82.48 & \meanvalue{83.34}
     & 87.87                                                        & 69.60 & 85.53 & 77.80 & 72.12             & 76.69 & 74.47 & 89.14 & \meanvalue{79.15} \\

    {\tt mDice $\uparrow$}
     & 87.18                                                        & 80.76 & 96.88 & 96.99 & \meanvalue{90.45}
     & 80.85                                                        & 92.94 & 85.76 & 93.19 & 67.04             & 66.57 & 71.31 & 82.63 & \meanvalue{80.04}
     & 84.71                                                        & 52.03 & 77.72 & 65.92 & 49.48             & 61.56 & 58.68 & 83.02 & \meanvalue{66.64} \\

    \bottomrule
\end{tabular}

    \end{adjustbox}
\end{table*}

\section{Comprehensive Quantitative Results}
\label{appendix:detailed_results}
In this section, we provide a comprehensive performance comparison between ConceptSeg-R1 and existing  methods. 

\textit{\textbf{Metrics:}} Beyond the metrics discussed in the main text, we introduce four additional evaluation metrics: Mean Absolute Error ({\tt MAE}), Balance Error Rate ({\tt BER}), S-measure ({\tt $S_m$}), and Mean Dice ({\tt mDice}). These metrics validate the superior performance of ConceptSeg-R1 across multiple dimensions, including pixel-level precision, error distribution, and structural similarity, where higher values are better for {\tt $S_m$} and {\tt mDice}, while lower values are preferred for {\tt MAE} and {\tt BER}. 
All results are reported in percentage (\%).

\textit{\textbf{Results:}} 
To evaluate model robustness across the cognitive spectrum, we report comprehensive results for the CI, CD, and CR concepts, as shown in \cref{tab:quantitative_full}. 
We provide a granular per-dataset breakdown for existing methods and our ConceptSeg-R1, complemented by the mean performance calculated for each concept category at the dataset level. This extensive reporting provides a transparent view of how different architectures handle specific domain challenges, e.g., long-tail distributions in CI or complex spatial constraints in CD.
The results consistently demonstrate that ConceptSeg-R1 outpaces representative models like LENS, Seg-Zero, and SAM3-Agent variants across most benchmarks. The rationale behind this superior performance lies in our  the Meta-GRPO reinforcement learning strategy and Concept Translation. Specifically, while conventional models often struggle with domain shifts, our framework utilizes rule induction to extract transferable knowledge from reference samples and deductive reasoning to maintain high precision in zero-shot scenarios. The inclusion of metrics like {\tt mDice} and {\tt $S_m$} further highlights that our model not only identifies the correct objects but also maintains superior structural integrity and boundary accuracy in complex environments.

\begin{table*}[t]
    \centering
\caption{Ablation study of reward components in ConceptSeg-R1-3B. Results are averaged over all concepts in CI, CD, and CR under all evaluation metrics.}
    \label{tab:app_reward_ablation}    
    \begin{adjustbox}{width=\linewidth}
        
\begin{tabular}{lcccccccccccccccccc}
\multirow{2}{*}[-2.5ex]{\textbf{Method}}& \multicolumn{6}{c}{\textit{CI Concepts}} & \multicolumn{6}{c}{\textit{CD Concepts}} & \multicolumn{6}{c}{\textit{CR Concepts}} \\
\cmidrule(lr){2-7} \cmidrule(lr){8-13} \cmidrule(lr){14-19}
 & {\tt MAE $\downarrow$} &{\tt BER $\downarrow$} & {\tt $F_{\beta}^{\omega} \uparrow$} & {\tt $S_m \uparrow$} &{\tt  mIoU $\uparrow$} & {\tt  mDice $\uparrow$} & {\tt  MAE $\downarrow$} & {\tt  BER$\downarrow$} & {\tt   $F_{\beta}^{\omega} \uparrow$} & {\tt  $S_m \uparrow$} & {\tt  mIoU $\uparrow$} & {\tt  mDice $\uparrow$} & {\tt   MAE $\downarrow$} & {\tt  BER $\downarrow$} & {\tt   $F_{\beta}^{\omega} \uparrow$} & {\tt  $S_m \uparrow$} &{\tt   mIoU $\uparrow$} & {\tt  mDice $\uparrow$} \\
\toprule
w/o box \& meta  & 1.56 & 4.78 & 89.81 & 89.29 & 91.70 & 90.33 & 7.43 & 11.22 & 75.85 & 80.94 & 80.67 & 77.56 & 9.44 & 18.65 & 47.44 & 67.70 & 67.69 & 51.05 \\
w/o mask $R_{mask}$ & 1.55 & 4.69 & 89.82 & 89.31 & 91.73 & 90.38 & 6.73 & \textbf{10.89} & \textbf{76.82} & 81.74 & 81.48 & \textbf{78.57} & 6.91 & 17.56 & 54.05 & 71.45 & 71.84 & 56.68 \\
\rowcolor{Light}
All Reward & \textbf{1.33} & \textbf{4.29} & \textbf{89.87} & \textbf{89.76} & \textbf{92.00} & \textbf{90.56} & \textbf{6.15} & 11.20 & 76.55 & \textbf{82.06} & \textbf{81.77} & 77.98 & \textbf{5.47} & \textbf{17.52} & \textbf{59.15} & \textbf{74.43} & \textbf{75.18} & \textbf{60.72} \\
\bottomrule
\end{tabular}

    \end{adjustbox} 
\end{table*}

\section{Reward Configuration Evaluation}
\label{appendix:Reward}

We conduct reward ablation experiments to analyze the contribution of different reward components in ConceptSeg-R1-3B, including mask reward, box reward, and meta reasoning reward. 
As shown in \cref{tab:app_reward_ablation}, removing either the box \& meta reward or the mask reward consistently degrades performance across CI, CD, and CR concepts. 
Among all settings, the full reward formulation achieves the best overall performance on nearly all metrics. 
In particular, removing the box and meta reward leads to a significant performance drop on CR concepts, indicating that structural grounding and reasoning-aware supervision are critical for complex reasoning segmentation. 
Meanwhile, removing the mask reward mainly affects CI and CD concepts, demonstrating the importance of pixel-level spatial supervision for perception-oriented segmentation. 
These results suggest that the different reward components provide complementary optimization signals, jointly improving spatial accuracy, structural consistency, and reasoning correctness in generalized concept segmentation.

\begin{table*}[t]
    \centering
    \caption{
    Ablation study of Shortcut Router between SAM 3 and the reasoning branch in ConceptSeg-R1-3B.
    The routing rate denotes the proportion of samples directly processed by SAM 3.}
    \label{tab:app_sam3_router_ablation}    
    \begin{adjustbox}{width=\linewidth}
        \begin{tabular}{lrcccrcccr}
                                                      &
                                                      &
    \multicolumn{4}{c}{$F_{\beta}^{\omega} \uparrow$} &
    \multicolumn{4}{c}{\tt mIoU $\uparrow$}                                                                                                                                                                                                        \\
    \cmidrule(lr){3-6} \cmidrule(lr){7-10}
    \textbf{Concept}                                  & \textbf{Routing Rate (\%)} & SAM 3 Only & Full Reasoning & Adaptive Reasoning & $\Delta$ Gains             & SAM 3 Only & Full Reasoning & Adaptive Reasoning & $\Delta$ Gains             \\
    \toprule

  \rowcolor{mygray}
    \multicolumn{10}{c}{\textit{\textbf{CI Concepts}}}                                                                                                                                                                                             \\
    \midrule
    Living Classes                                    & 97.60                      & 86.28      & 72.45          & 86.17              & \textbf{\increase{+13.72}} & 88.66      & 80.54          & 88.55              & \textbf{\increase{+08.01}} \\
    Artifact Classes                                  & 96.10                      & 80.11      & 61.91          & 79.93              & \textbf{\increase{+18.02}} & 85.40      & 75.56          & 85.20              & \textbf{\increase{+09.64}} \\
    Fine-grained Classes                              & 86.50                      & 96.03      & 96.65          & 96.92              & \textbf{\increase{+00.27}} & 96.74      & 97.24          & 97.38              & \textbf{\increase{+00.14}} \\
    Ultra Rare Classes                                & 7.70                       & 95.72      & 95.58          & 96.45              & \textbf{\increase{+00.87}} & 96.47      & 96.37          & 96.87              & \textbf{\increase{+00.50}} \\

\rowcolor{mygray}
    \multicolumn{10}{c}{\textit{\textbf{CD Concepts}}}                                                                                                                                                                                             \\
    \midrule
    Optical Property (Shadow)                         & 99.50                      & 79.67      & 43.45          & 79.82              & \textbf{\increase{+36.37}} & 81.49      & 58.50          & 81.56              & \textbf{\increase{+23.06}} \\
    Optical Property (Transparency)                   & 99.90                      & 91.46      & 69.70          & 91.57              & \textbf{\increase{+21.87}} & 91.34      & 74.37          & 91.37              & \textbf{\increase{+17.00}} \\
    Camouflage                                        & 8.40                       & 51.28      & 83.94          & 83.65              & \textbf{\decrease{-00.29}} & 61.43      & 87.96          & 87.65              & \textbf{\decrease{-00.31}} \\
    Saliency                                          & 0.00                       & 38.06      & 88.96          & 88.96              & \meanvalue{0.00}           & 58.97      & 90.47          & 90.47              & \meanvalue{0.00}           \\
    Industrial Anomaly                                & 0.00                       & 51.48      & 61.52          & 61.52              & \meanvalue{0.00}           & 66.17      & 70.97          & 70.97              & \meanvalue{0.00}           \\
    Polyp Lesion                                      & 0.00                       & 17.46      & 58.96          & 58.96              & \meanvalue{0.00}           & 44.91      & 74.31          & 74.31              & \meanvalue{0.00}           \\
    Breast Lesion                                     & 0.70                       & 49.57      & 66.90          & 66.75              & \textbf{\decrease{-00.15}} & 65.53      & 76.52          & 76.33              & \textbf{\decrease{-00.19}} \\
    Skin Lesion                                       & 0.00                       & 41.86      & 81.14          & 81.14              & \meanvalue{0.00}           & 36.31      & 81.50          & 81.50              & \meanvalue{0.00}           \\

    \rowcolor{mygray}
    \multicolumn{10}{c}{\textit{\textbf{CR Concepts}}}                                                                                                                                                                                             \\
    \midrule
    Common Consistency                                & 0.00                       & --         & 80.11          & 80.11              & \meanvalue{0.00}           & --         & 85.77          & 85.77              & \meanvalue{0.00}           \\
    Correspondence Consistency                        & 0.00                       & --         & 34.59          & 34.59              & \meanvalue{0.00}           & --         & 59.69          & 59.69              & \meanvalue{0.00}           \\
    Reference Consistency                             & 0.00                       & --         & 77.07          & 77.07              & \meanvalue{0.00}           & --         & 85.57          & 85.57              & \meanvalue{0.00}           \\
    Static Difference                                 & 0.00                       & --         & 49.68          & 49.68              & \meanvalue{0.00}           & --         & 68.21          & 68.21              & \meanvalue{0.00}           \\
    View Difference                                   & 0.00                       & --         & 55.78          & 55.78              & \meanvalue{0.00}           & --         & 75.41          & 75.41              & \meanvalue{0.00}           \\
    Logical Rationality                               & 0.00                       & --         & 46.57          & 46.57              & \meanvalue{0.00}           & --         & 68.49          & 68.49              & \meanvalue{0.00}           \\
    Cross-View Spatio-temporality                     & 0.00                       & --         & 49.90          & 49.90              & \meanvalue{0.00}           & --         & 70.24          & 70.24              & \meanvalue{0.00}           \\
    Cross-Frame Spatio-temporality                    & 0.00                       & --         & 79.51          & 79.51              & \meanvalue{0.00}           & --         & 88.05          & 88.05              & \meanvalue{0.00}           \\
    \bottomrule
\end{tabular}

    \end{adjustbox} 
\end{table*}

\section{Routing Strategy Evaluation}
\label{appendix:Router}

In \cref{tab:app_sam3_router_ablation}, we compare three routing strategies: 
always using the direct SAM 3 segmentation pathway (SAM 3 only), 
always executing the full reasoning pipeline (Full Reasoning), 
and the proposed adaptive routing strategy (Adaptive Reasoning).
The results demonstrate that adaptive routing effectively balances efficiency and reasoning capability across different concept regimes. 
On CI concepts, where strong reasoning is unnecessary, the adaptive strategy achieves performance comparable to the direct SAM 3 pathway while consistently outperforming the full reasoning pipeline, indicating that the router preserves the native segmentation capability of SAM 3 without introducing unnecessary reasoning overhead. 
More importantly, the routing behavior adapts to task complexity within the CI regime. 
For relatively simple categories such as living and artifact classes, the routing rate remains above 95\%, and the adaptive performance matches the direct SAM 3 results. 
In contrast, for more challenging categories such as fine-grained and ultra rare classes, an increasing portion of samples is routed to the reasoning branch. 
Specifically, on fine-grained classes, both ``Full Reasoning'' and ``Adaptive Reasoning'' outperform the ``SAM 3 Only'', indicating the benefit of reasoning for subtle visual distinctions. 
On ultra rare classes, ``Full Reasoning'' alone slightly underperforms ``SAM 3 Only'', whereas the proposed ``Adaptive Reasoning'' consistently achieves the best performance. 
These results demonstrate that the router dynamically balances perception and reasoning according to task difficulty.
On CD concepts, the adaptive strategy yields substantial improvements over always reasoning, particularly in shadow and transparency concepts, where gains of up to +0.23 {\tt mIoU} are observed. 
A closer inspection reveals distinct routing patterns across different CD concept understanding tasks. 
For shadow and transparency concepts, the direct SAM 3 pathway already achieves strong performance and significantly outperforms the ``Full Reasoning'', indicating that these tasks primarily depend on accurate visual perception rather than complex reasoning. 
In contrast, for other CD concepts, the ``Full Reasoning''  consistently surpasses the ``SAM 3 Only'', demonstrating the necessity of contextual reasoning in these scenarios. 
Importantly, the proposed ``Adaptive Reasoning'' strategy learns to route samples accordingly: 
the routing rate approaches 100\% for shadow and transparency concepts, while remaining near zero for other CD concepts. 
This behavior confirms that the router performs precise sample-level decision, activating reasoning only when beneficial and preserving direct perception when sufficient.  
On CR concepts, the routing rate consistently drops to zero, meaning that all samples are processed through the reasoning branch. 
This result validates that the router correctly identifies reasoning-intensive tasks and fully engages the reasoning pipeline when multi-step inference is necessary. 
Overall, these findings verify that the Shortcut Router enables dynamic allocation of computational resources, preserving efficiency on simple cases while maintaining strong reasoning performance on complex concept understanding tasks.

\begin{figure*}[!t]
  \centering
  \includegraphics[width=1\linewidth]{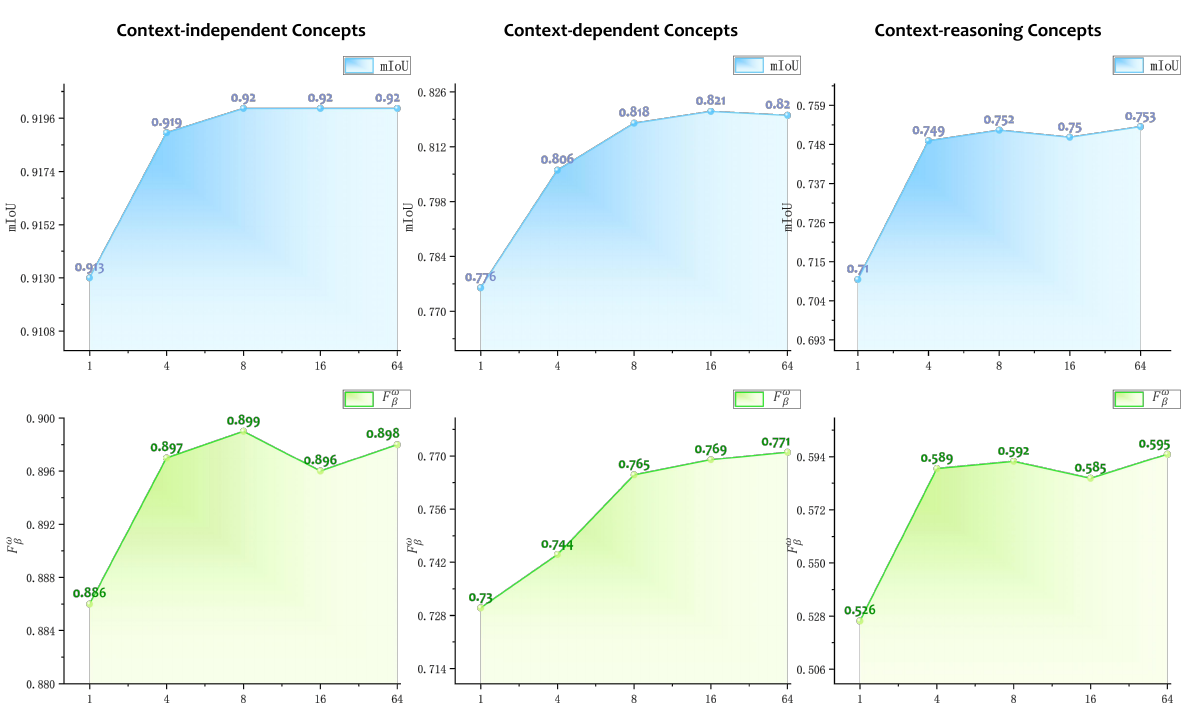}
  \caption{Impact of the concept query length ($L_2$) on concept segmentation benchmarks. Results for \texttt{mIoU} and $F_{\beta}^{\omega}$ are reported as dataset-level averages.}
  \label{fig:concept_query_length_ablation}
\end{figure*}

\section{Concept Query Sensitivity}
\label{appendix:Translation}

We analyze the impact of varying the number of learnable concept queries (i.e., $L_2$ in $\mathbf{C}\in\mathbb{R}^{L_2\times C}$ as stated in~\cref{sec:concept_translation,equ:crossattn_in_ctm}), to assess the module's sensitivity to contextual information, as illustrated in \cref{fig:concept_query_length_ablation}.
Our analysis reveals two key insights:
\begin{itemize}[leftmargin=*,itemsep=0em,topsep=0em,parsep=0em]
    \item \textbf{Semantic Bottleneck}. When $L_2=1$, performance across all concept segmentation benchmarks is markedly suboptimal. This degradation stems from a semantic bottleneck between reasoning and segmentation: a single query token is insufficient to encapsulate the high-dimensional reasoning outputs from the MLLM, thereby failing to provide SAM 3 with effective guidance for pixel-level localization.
    \item \textbf{Robustness and Convergence}. As $L_2$ increases, we observe a consistent performance gain that converges at $L_2=8$. This trend suggests that the model is highly robust once the query length exceeds this threshold, as $L_2 \ge 8$ effectively alleviates the bottleneck by providing sufficient semantic bandwidth to convey refined prompt for SAM 3. Consequently, to maintain an optimal trade-off between computational efficiency and segmentation accuracy, we fix $L_2=8$ for our final configuration.
\end{itemize}

\begin{table*}[t]
    \centering
\caption{Evaluation of ConceptSeg-R1 under different randomly sampled reference prompts (mean $\pm$ std over three runs).}
    \label{tab:app_random_3run}    
    \begin{adjustbox}{width=0.8\linewidth}
        \begin{tabular}{llcccccc}

\textbf{Concept} & {\tt MAE} $\downarrow$ & {\tt BER} $\downarrow$ & {\tt $F_{\beta}^{\omega}$} $\uparrow$ & {\tt $S_m$} $\uparrow$ & {\tt mIoU} $\uparrow$ & {\tt mDice} $\uparrow$ \\
\toprule
CI & 1.35$\pm$0.02  & 4.38$\pm$0.12  & 89.76$\pm$0.16  & 89.55$\pm$0.29  & 91.89$\pm$0.16  & 90.44$\pm$0.16  \\
CD & 6.06$\pm$0.12  & 11.19$\pm$0.01  & 76.30$\pm$0.35  & 81.94$\pm$0.18  & 81.75$\pm$0.03  & 77.86$\pm$0.17  \\
CR & 5.44$\pm$0.05  & 17.31$\pm$0.29  & 59.14$\pm$0.01  & 74.36$\pm$0.10  & 75.17$\pm$0.02  & 60.69$\pm$0.05  \\
\bottomrule
\end{tabular}
    \end{adjustbox} 
\end{table*}

\begin{table*}[t]
    \centering
\caption{Ablation experiments of different reference configurations ($1\times1$ vs. $2\times2$).}
    \label{tab:app_reference_number_ablation}    
    \begin{adjustbox}{width=0.6\linewidth}
        
\begin{tabular}{cccccccc}

\textbf{Concept} & \textbf{Reference}& {\tt MAE} $\downarrow$ & {\tt BER} $\downarrow$ & {\tt $F_{\beta}^{\omega}$} $\uparrow$ & {\tt $S_m$} $\uparrow$ & {\tt mIoU} $\uparrow$ & {\tt mDice} $\uparrow$  \\
\toprule
\multirow{3}{*}{CI} 
 & 1$\times$1 & 1.40 & 4.53 & 89.54 & 89.27 & 91.70 & 90.24 \\
 & 2$\times$2 & 1.33 & 4.29 & 89.87 & 89.76 & 92.00 & 90.56 \\
 & $\Delta$ Gains & \increase{+0.07} & \increase{+0.24} & \decrease{-0.33} & \decrease{-0.49} & \decrease{-0.30} & \decrease{-0.32} \\
\midrule
\multirow{3}{*}{CD} 
 & 1$\times$1 & 7.22 & 11.23 & 75.01 & 80.77 & 80.58 & 76.92 \\
 & 2$\times$2 & 6.15 & 11.20 & 76.55 & 82.06 & 81.77 & 77.98 \\
 & $\Delta$ Gains & \increase{+1.07} & \increase{+0.03} & \decrease{-1.54} & \decrease{-1.29} & \decrease{-1.19} & \decrease{-1.06} \\
\midrule
CR & Default & 5.47 & 17.52 & 59.15 & 74.43 & 75.18 & 60.72 \\
\bottomrule
\end{tabular}
    \end{adjustbox} 
\end{table*} 

\section{Reference Robustness and Configuration}
\label{appendix:training_random_prompt}
We evaluate the robustness of ConceptSeg-R1 under different reference prompts and analyzes the impact of reference configurations. 
\begin{itemize}[leftmargin=*,itemsep=0em,topsep=0em,parsep=0em]
\item \textbf{Random Reference Stability.}
We first evaluate the stability of ConceptSeg-R1 using randomly sampled reference prompts across three independent runs with different random seeds. 
As shown in Tab.~\ref{tab:app_random_3run}, the model achieves highly consistent performance across CI, CD, and CR concepts, with all standard deviations remaining very small under different metrics. 
Among the three concept regimes, CI concepts exhibit the highest stability, while CD and CR concepts also maintain robust performance despite involving more complex contextual dependency and reasoning processes. 
During training, the reference prompts are randomly composed from different support examples, forcing the model to induce transferable task rules rather than overfitting to fixed prompt layouts or specific instances. 
This randomized prompt composition improves robustness to diverse reference configurations and enhances generalization under varying prompt distributions. 
These results demonstrate that ConceptSeg-R1 is insensitive to specific reference selections and generalizes reliably across diverse prompt compositions.

\item \textbf{Reference Configuration Ablation.}
We further compare different reference configurations by evaluating a $1\times1$ single-reference setting and a $2\times2$ mosaic-reference setting. 
As reported in Tab.~\ref{tab:app_reference_number_ablation}, the $2\times2$ configuration consistently outperforms the $1\times1$ setting on both CI and CD concepts across nearly all metrics. 
The performance improvement is especially significant on CD concepts, indicating that multiple references provide richer contextual diversity and stronger cross-image consistency for task rule induction. 
In contrast, single-reference prompts are more susceptible to ambiguity and stochastic bias, making it harder for the model to infer reliable task-level concepts. 
For CR concepts, we adopt the default reference configuration due to the stronger dependence on reasoning-oriented prompt composition.
\end{itemize}

\begin{figure*}[!t]
  \centering 
  \includegraphics[width=1\linewidth]{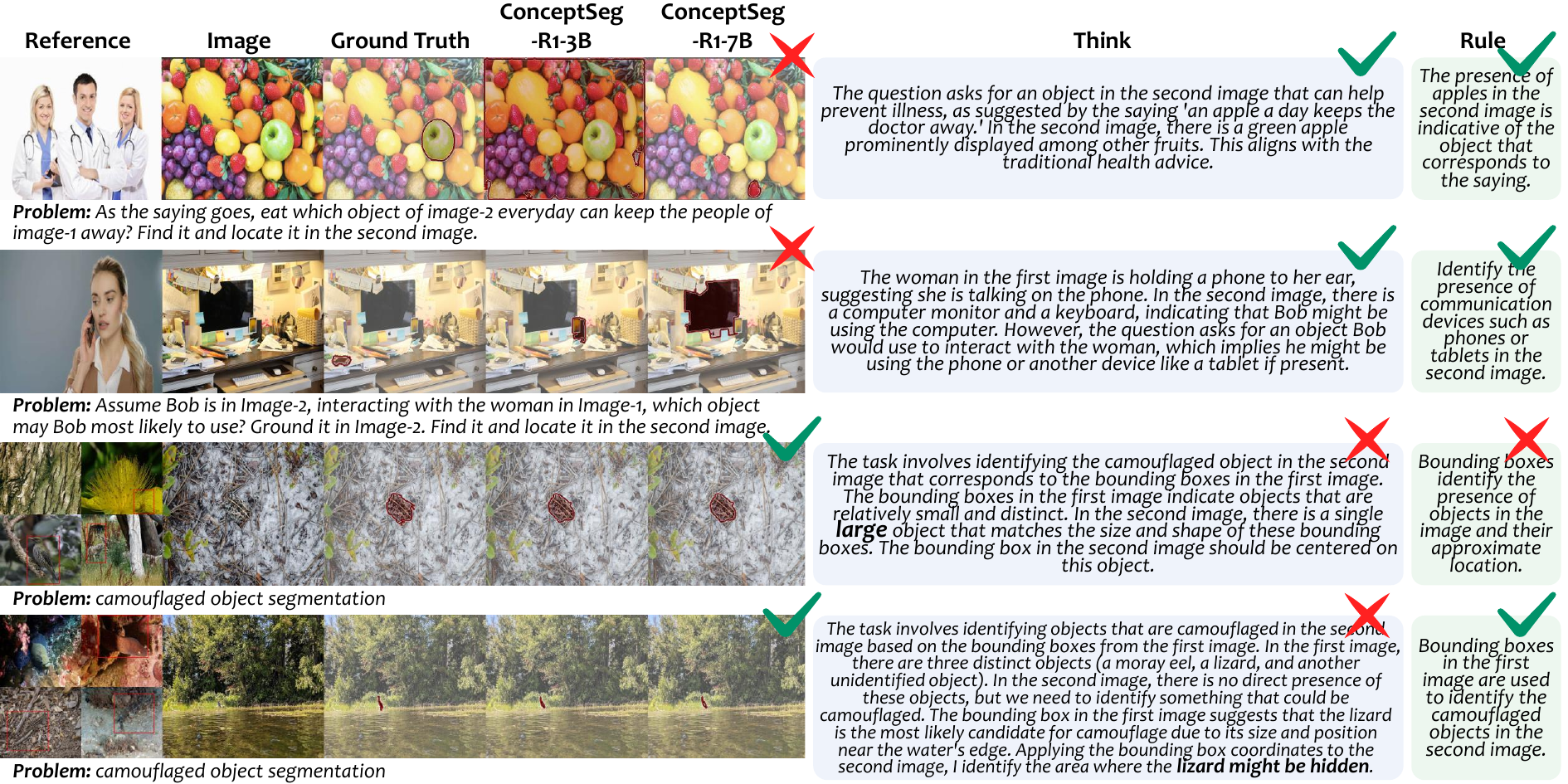}
\caption{Visualization of the correctness relationships among segmentation prediction, reasoning trajectory ({\tt Think}), and induced rule ({\tt Rule}) in reasoning segmentation.}
  \label{fig:App_Reasoning_Correct}
\end{figure*}

\section{Beyond Mask Accuracy: Decomposing Reasoning Segmentation}

Existing reasoning-oriented segmentation benchmarks primarily evaluate the final segmentation mask, typically using overlap-based metrics such as {\tt mIoU} or {\tt cIoU}. 
However, a correct mask does not necessarily imply correct reasoning. 
A model may produce visually plausible predictions while relying on incorrect rules, spurious correlations, or shortcut reasoning paths. 
Therefore, evaluating only the final segmentation result is insufficient for understanding the true reasoning capability of concept segmentation systems.

To further analyze this issue, we provide a qualitative decomposition of the reasoning process in Fig.~\ref{fig:App_Reasoning_Correct}. 
Instead of evaluating only the final prediction, we explicitly examine the consistency among three components: the inferred reasoning trajectory ({\tt Think}),  the induced rule ({\tt Rule}), and 
 the final segmentation result ({\tt Prediction}). 
Each component may independently succeed or fail, leading to different forms of reasoning correctness and failure cases.
For example, a model may generate the correct segmentation mask while relying on incorrect reasoning logic, indicating shortcut learning rather than genuine concept understanding. 
Conversely, a model may induce semantically correct rules but fail during the final segmentation execution due to inaccurate grounding or imperfect mask decoding. 
These observations suggest that reasoning segmentation should be evaluated as a multi-stage process involving rule induction, reasoning consistency, grounding correctness, and final mask quality, rather than a single end-point prediction task.
We hope this analysis can motivate future research toward more comprehensive evaluation protocols for reasoning segmentation, including intermediate reasoning supervision, rule consistency verification, and causal grounding assessment. 

\section{Qualitative Visualization}
\label{appendix:Visualization}
\begin{figure*}[t]
  \centering
  \includegraphics[width=\linewidth]{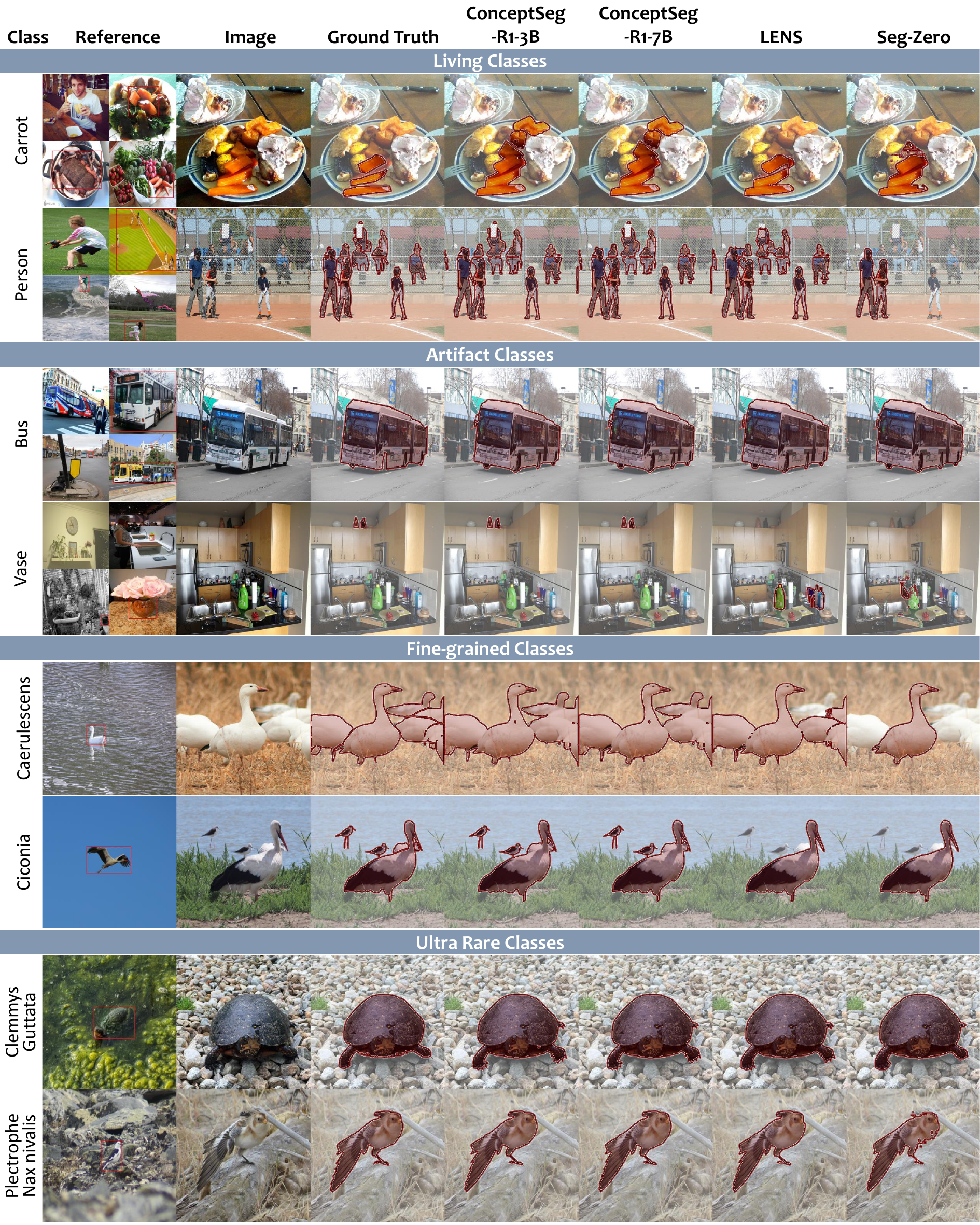}
  \caption{Qualitative comparison on diverse CI concepts.}
  \label{fig:app_ci_classes_visual}
\end{figure*}

\begin{figure*}[!t]
  \centering
  \includegraphics[width=0.9\linewidth]{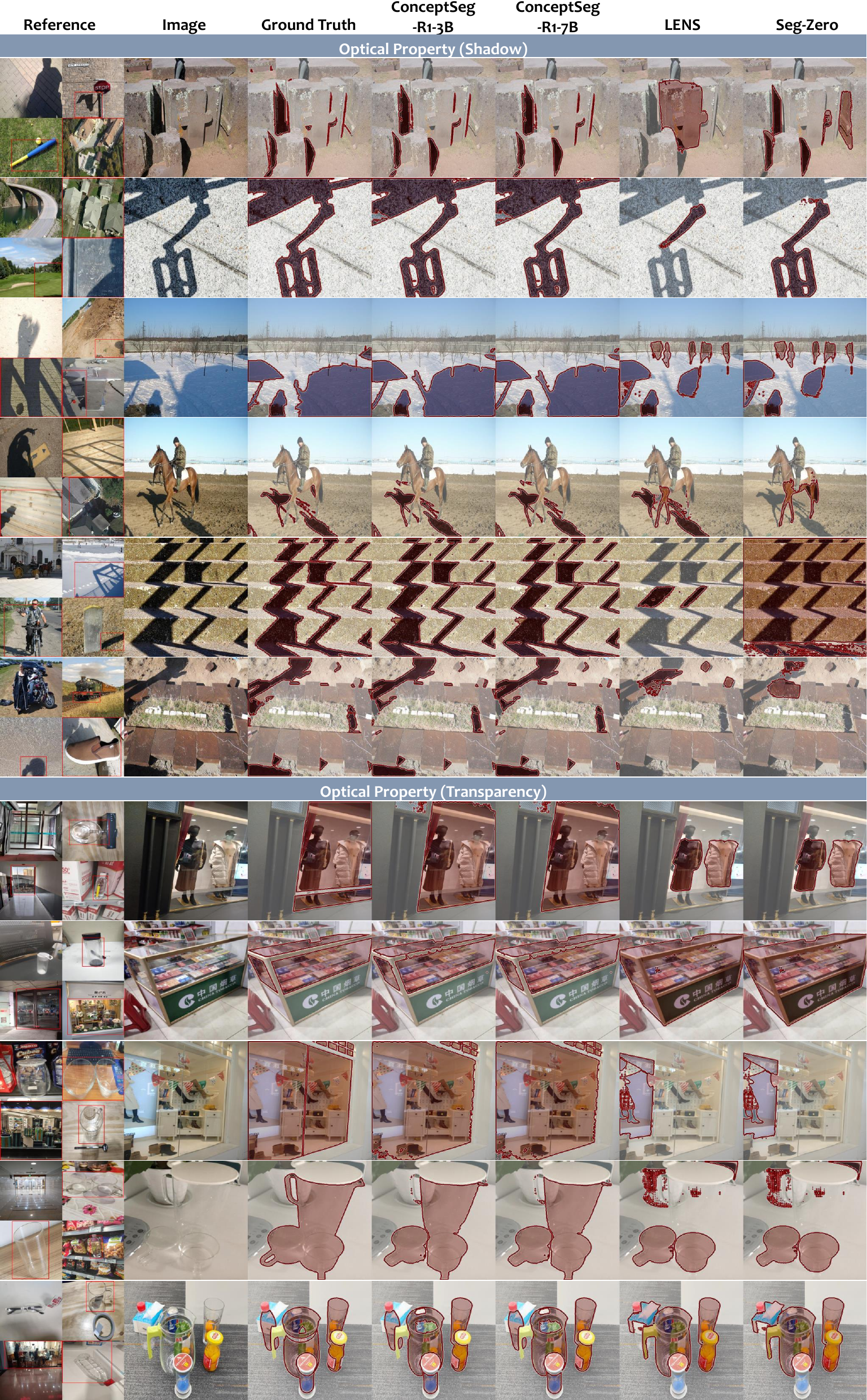}
  \caption{Qualitative comparison on optical property concepts.}
  \label{fig:app_optical_visual}
\end{figure*}

\begin{figure*}[!t]
  \centering
  \includegraphics[width=1\linewidth]{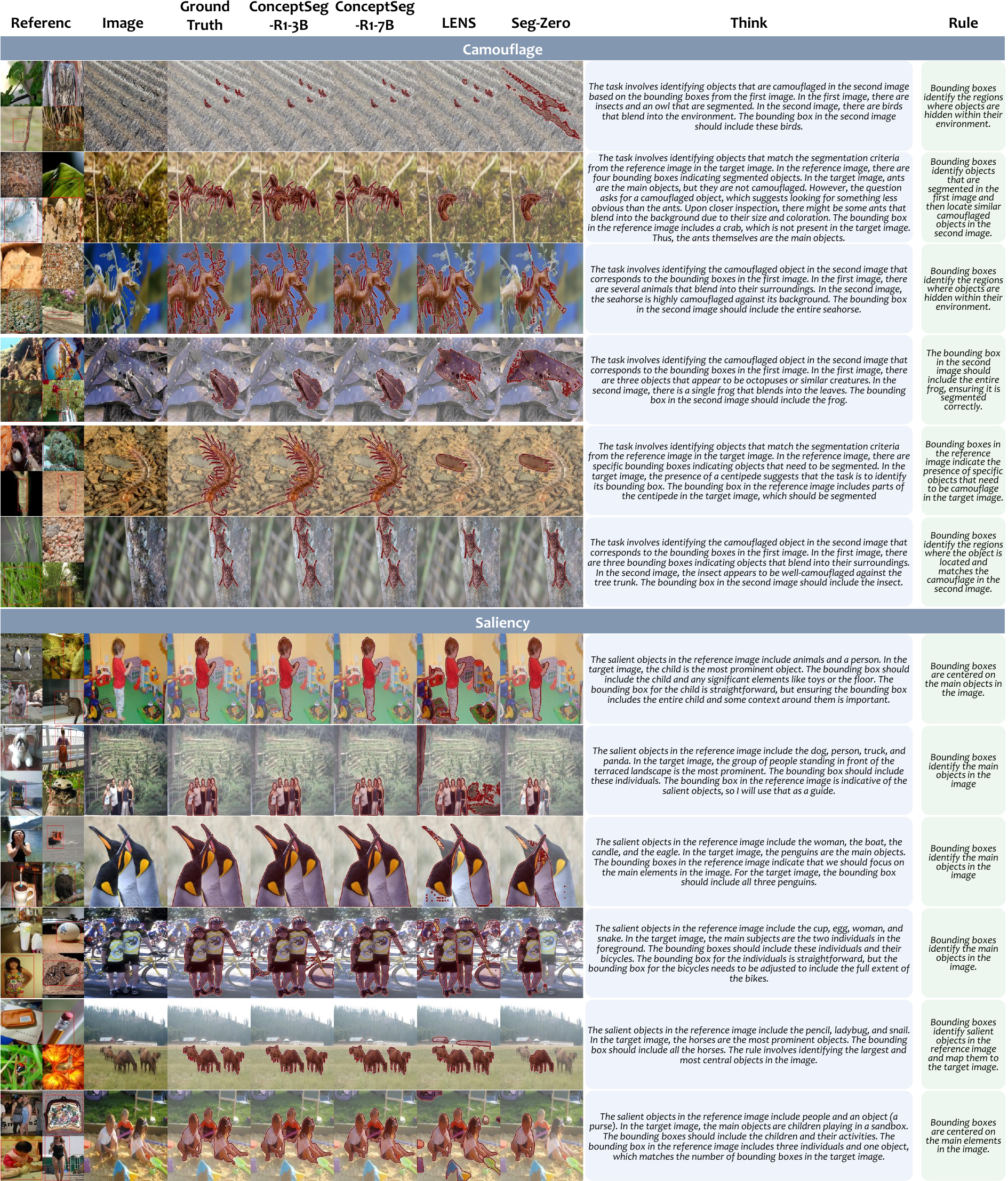}
  \caption{Qualitative comparison on camouflage and saliency concepts.}
  \label{fig:app_camu_salient_visual}
\end{figure*}

\begin{figure*}[!t]
  \centering
  \includegraphics[width=1\linewidth]{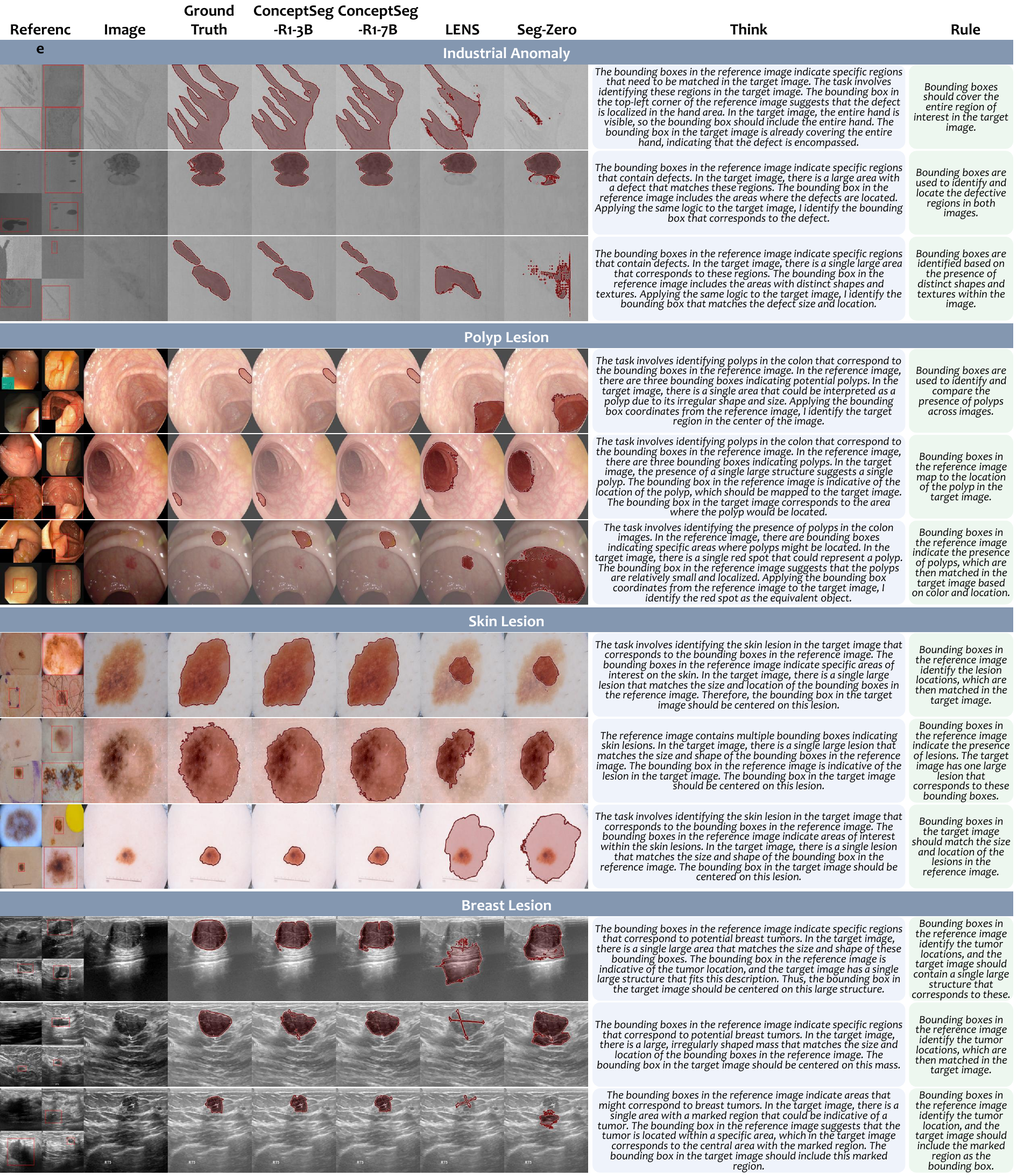}
  \caption{Qualitative comparison on industrial anomaly and medical lesion concepts.}
  \label{fig:appendix_Industrial_medical_Concept_visual}
\end{figure*}

\begin{figure*}[!t]
  \centering 
  \includegraphics[width=1\linewidth]{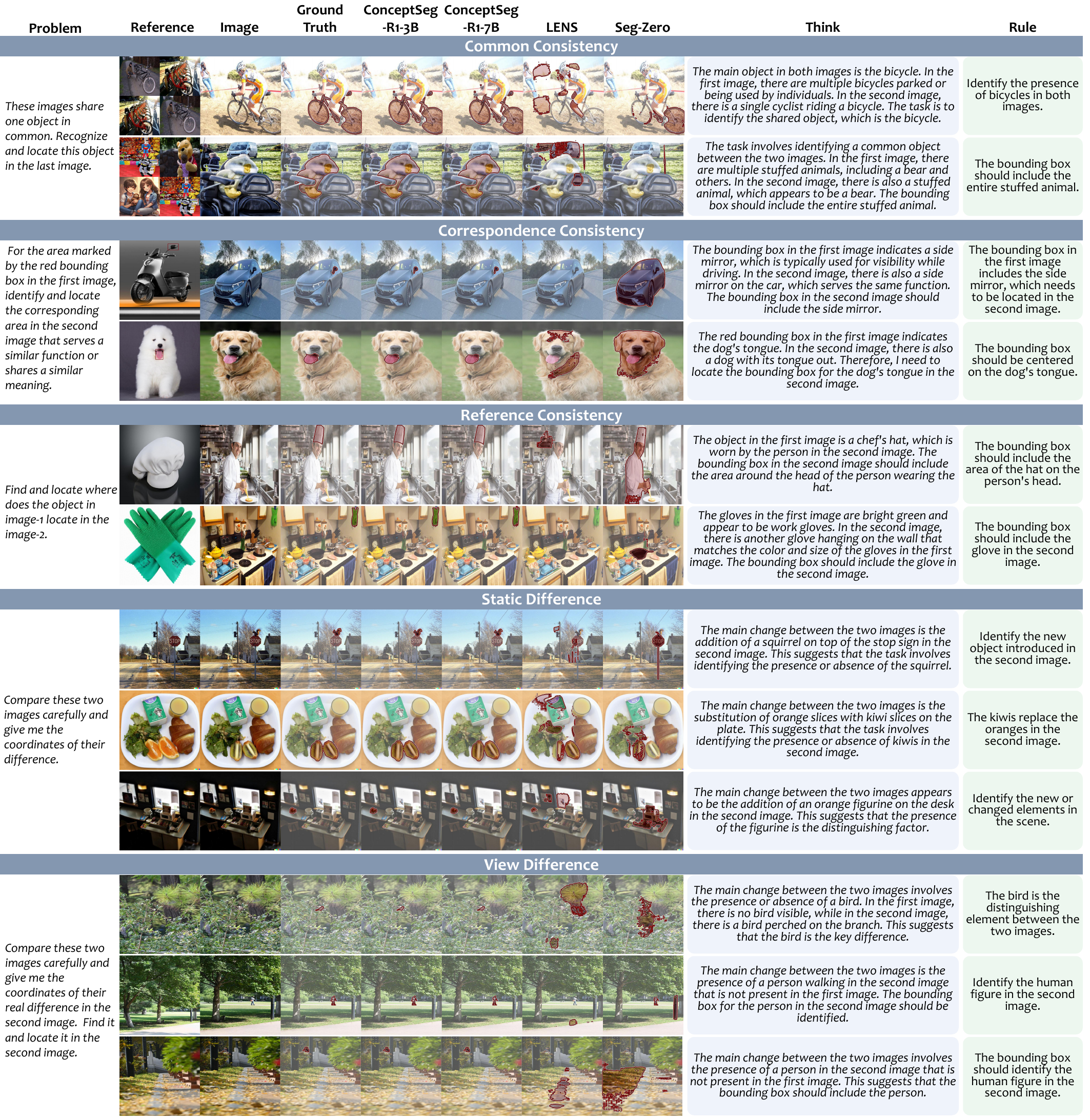}
  \caption{Qualitative comparison on consistency and difference reasoning concepts.}
  \label{fig:app_consistency_difference_visual}
\end{figure*}

\begin{figure*}[!t]
  \centering
  \includegraphics[width=1\linewidth]{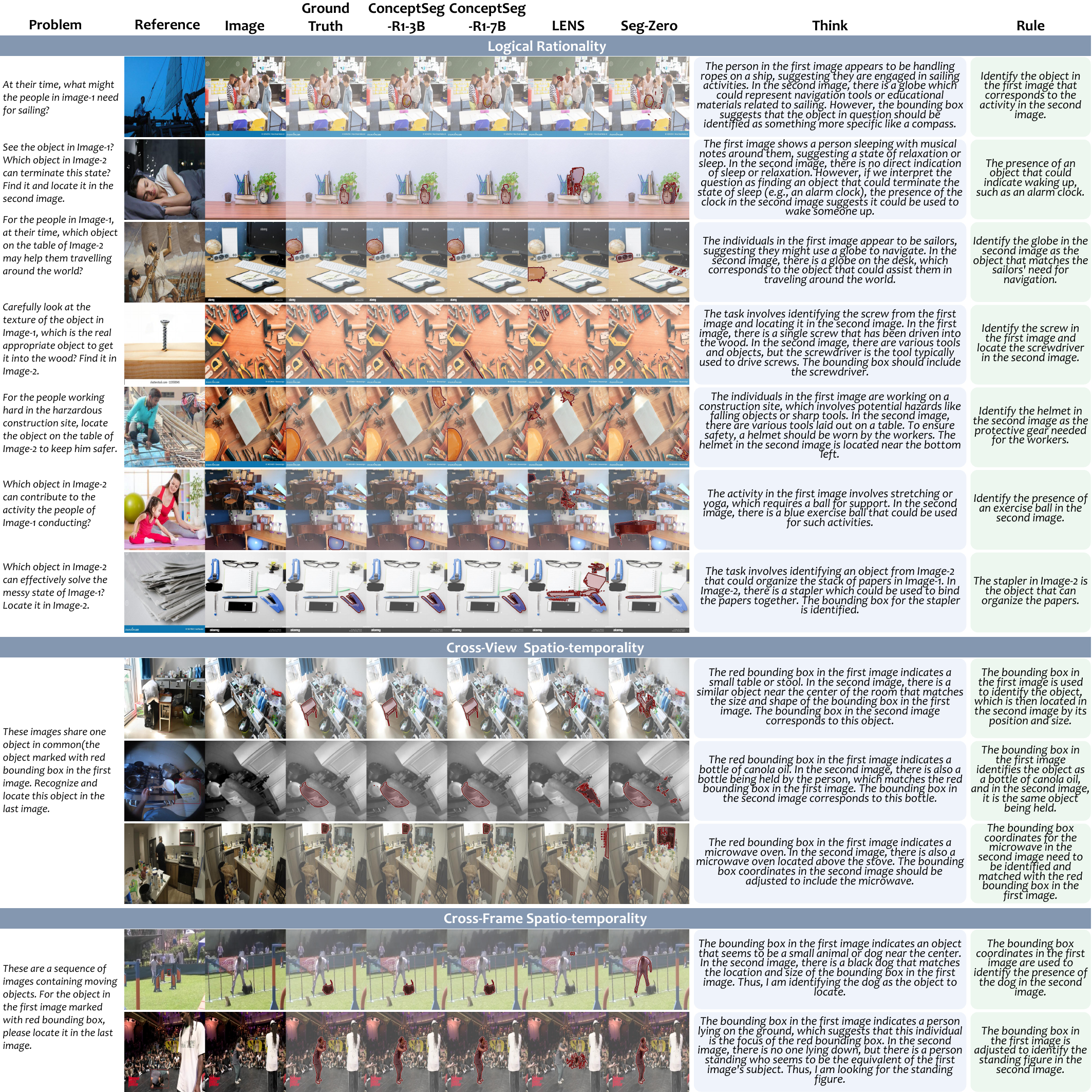}
  \caption{Qualitative comparison on logical rationality and spatio-temporality reasoning concepts.}
  \label{fig:app_logic_spatio_temporal}
\end{figure*}

We provide additional qualitative comparisons across CI, CD, and CR concepts, as shown in~\cref{fig:app_ci_classes_visual,fig:app_optical_visual,fig:app_camu_salient_visual,fig:app_consistency_difference_visual,fig:appendix_Industrial_medical_Concept_visual,fig:app_logic_spatio_temporal}. 

\section{Limitations and Future Works}

As an initial step toward segmenting any concept, our ConceptSeg-R1 still presents several limitations and opportunities for future improvement.
\noindent\textit{\textbf{\uppercase\expandafter{\romannumeral1})  Dependence on Representative Support Examples.}}
The effectiveness of rule induction relies on the quality and representativeness of the provided support examples. When support instances are highly ambiguous, visually noisy, or weakly aligned with the target concept, the inferred rule may be incomplete or suboptimal, potentially affecting downstream segmentation accuracy. 
Future work may address this limitation by incorporating automated support selection mechanisms, active example filtering, or confidence-aware retrieval strategies to improve the reliability of rule induction under challenging conditions.
\noindent\textit{\textbf{\uppercase\expandafter{\romannumeral2}) Scalability to Extremely Long Reasoning Chains.}}
Current reasoning procedures are performed within a fixed context window and are primarily optimized for short to moderate reasoning sequences. Tasks requiring very long reasoning chains, complex temporal dependencies, or hierarchical multi-step planning may exceed the effective reasoning capacity of the current architecture. 
Future research could explore hierarchical reasoning structures, memory-augmented reasoning modules, or external reasoning buffers to support more scalable reasoning workflows.
\noindent\textit{\textbf{\uppercase\expandafter{\romannumeral3}) Sensitivity to Routing Decisions in Dynamic Inference.}}
The shortcut routing mechanism improves efficiency by dynamically selecting between lightweight and full reasoning pipelines. However, incorrect routing decisions may occasionally lead to suboptimal performance, particularly when the input complexity lies near the boundary between concept categories. 
Future work may improve routing robustness through uncertainty-aware routing policies, adaptive threshold learning, or lightweight verification stages that reduce the impact of routing errors.
\noindent\textit{\textbf{\uppercase\expandafter{\romannumeral4}) Computational Overhead in Reasoning-Enhanced Segmentation.}}
Although ConceptSeg-R1 maintains practical inference latency, reasoning-enhanced segmentation generally introduces additional computational cost compared to purely feed-forward segmentation models. This overhead may become more pronounced in large-scale deployment scenarios or real-time applications with strict latency constraints. 
Future optimization efforts may focus on model compression, dynamic reasoning pruning, or hardware-aware scheduling to further improve computational efficiency while preserving reasoning capability.

\clearpage

\clearpage

\bibliography{main}

\end{document}